\documentclass[10pt,journal,compsoc]{IEEEtran}

\usepackage{hyperref}
\usepackage{url}
\usepackage{float}
\usepackage{enumitem}
\usepackage{amsmath,amsfonts,amssymb}
\usepackage{epsfig}
\usepackage{subfigure}
\usepackage{graphicx}
\usepackage{textcomp}
\usepackage[normalem]{ulem}
\usepackage{epstopdf}
\usepackage{dsfont}
\usepackage{hhline}
\usepackage{booktabs}
\usepackage{color} 
\usepackage{float}
\usepackage{tcolorbox}
\usepackage{lipsum}
\tcbuselibrary{breakable}
\usepackage{multirow}
\usepackage{pbox}

\newcommand{\mt}[1]{\mathrm{#1}}

\newcommand{\iin}{\mathrm{in}}
\newcommand{\out}{\mathrm{out}}

\newcommand{\definemail}[2]{\newrobustcmd#1{\href{mailto:#2}{#2}}}
\definemail\authorone{amir\_ghasemian@fas.harvard.edu}
\definemail\authortwo{homahoss@isi.edu}
\definemail\authorthree{aaron.clauset@colorado.edu}
\newcommand{\amir}[1]{{\color{brown} #1}}

\ifCLASSOPTIONcompsoc
  \usepackage[nocompress]{cite}
\else
  \usepackage{cite}
\fi

\ifCLASSINFOpdf

\else

\fi

\begin{document}
\title{Evaluating Overfit and Underfit in Models of Network Community Structure}

\author{Amir~Ghasemian,~\IEEEmembership{}
        Homa~Hosseinmardi~\IEEEmembership{}
        and~Aaron~Clauset~\IEEEmembership{}
\IEEEcompsocitemizethanks{\IEEEcompsocthanksitem A. Ghasemian is with the 
Department of Statistics, Harvard University, Cambridge, MA, 02138.
\protect\\
E-mail: \authorone
\IEEEcompsocthanksitem H. Hosseinmardi is with the Information Sciences Institute, University of Southern California, Marina del Rey, CA, 90292.
\protect\\
Email: \authortwo
\IEEEcompsocthanksitem A. Clauset is an Associate Professor of Computer Science, University of Colorado, Boulder, CO, 80309, and in the BioFrontiers Institute, University of Colorado, Boulder, CO, 80305. Also he is an external Faculty at the Santa Fe Institute, 1399 Hyde Park Road, Santa Fe, NM, 87501.  
\protect\\
Email: \authorthree}
\thanks{}}

\markboth{}%
{Ghasemian \MakeLowercase{\textit{et al.}}: Bare Demo of IEEEtran.cls for Computer Society Journals}

\IEEEtitleabstractindextext{
\begin{abstract}
A common graph mining task is community detection, which seeks an unsupervised decomposition of a network into groups based on statistical regularities in network connectivity. Although many such algorithms exist, community detection's No Free Lunch theorem implies that no algorithm can be optimal across all inputs. However, little is known in practice about how different algorithms over or underfit to real networks, or how to reliably assess such behavior across algorithms. Here, we present a broad investigation of over and underfitting across 16 state-of-the-art community detection algorithms applied to a novel benchmark corpus of 572 structurally diverse real-world networks. We find that (i) algorithms vary widely in the number and composition of communities they find, given the same input; (ii) algorithms can be clustered into distinct high-level groups based on similarities of their outputs on real-world networks; (iii) algorithmic differences induce wide variation in accuracy on link-based learning tasks; and, (iv) no algorithm is always the best at such tasks across all inputs. Finally, we quantify each algorithm's overall tendency to over or underfit to network data using a theoretically principled diagnostic, and discuss the implications for future advances in community detection.
\end{abstract}

}

\maketitle

\IEEEdisplaynontitleabstractindextext
\IEEEpeerreviewmaketitle

\IEEEraisesectionheading{\section{Introduction}\label{sec:intro}}
\IEEEPARstart{N}{etworks} are an increasingly important and common kind of data, arising in social, technological, communication, and biological settings. One of the most common data mining tasks in network analysis and modeling is to coarse-grain the network, which is typically called community detection. This task is similar to clustering, in that we seek a lower-dimensional description of a network by identifying statistical regularities or patterns in connections among groups of nodes. Fundamentally, community detection searches for a partition of the nodes that optimizes an objective function of the induced clustering of the network.

Due to broad interest across disciplines in clustering networks, many approaches for community detection now exist~\cite{porter:etal:2010review,fortunato2010review,peixoto2017bayesian}, and these can be broadly categorized into either probabilistic methods or non-probabilistic methods. Graphical models, like the popular stochastic block model (SBM)~\cite{holland1983stochastic}, typically fall in the former category, while popular methods like modularity maximization~\cite{newman2004finding} fall in the latter. Across these general categories, methods can also be divided into roughly six groups: Bayesian and regularized likelihood approaches~\cite{hofman2008bayesian,yan2016bayesian,daudin2008mixture}, spectral and embedding techniques~\cite{Krzakala2013,saade2014spectral,le2015estimating}, modularity methods~\cite{newman2004finding}, information theoretic approaches such as Infomap~\cite{rosvall2008maps}, statistical hypothesis tests~\cite{wang2015likelihood}, and cross-validation methods~\cite{chen2014network,kawamoto2016cross}.

Despite great interest, however, there have been relatively few broad comparative studies or systematic evaluations of different methods in practical settings~\cite{hric:darst:fortunato:2014,le2015estimating,kawamoto2016comparative} and little is known about the degree to which different methods perform well on different classes of networks in practice. As a result, it is unclear which community detection algorithm should be applied to which kind of data or for which kind of downstream task, or how to decide which results are more or less useful when different algorithms produce different results on the same input~\cite{fortunato:hric:2016}.

This situation is worsened by two recently proved theorems for community detection~\cite{peel2017ground}. The No Free Lunch (NFL) theorem for community detection implies that no method can be optimal on all inputs, and hence every method must make a tradeoff between better performance on some kinds of inputs for worse performance on others. For example, an algorithm must choose a number of communities $k$ to describe a given network, and the way it makes this decision embodies an implicit tradeoff that could lead to overfitting (finding too many clusters) on some networks and underfitting (finding too few or the wrong clusters) on others. 

The ``no ground truth'' theorem states that there is no bijection between network structure and ``ground truth'' communities, which implies that no algorithm can always recover the correct ground truth on every network~\cite{peel2017ground}, even probabilistically. Together, these theorems have broad implications for measuring the efficacy of community detection algorithms. In the most popular evaluation scheme, a partition defined by node metadata or node labels is treated as if it were ``ground truth'', e.g., ethnicity in a high-school social network or cellular function in a protein interaction network, and accuracy on its recovery is compared across algorithms. However, the NFL and ``no ground truth'' theorems imply that such comparisons are misleading at best, as performance differences are confounded by implicit algorithmic tradeoffs across inputs~\cite{peel2017ground}. Hence, relatively little is known about how over- and under-fitting behavior varies by algorithm and input, past evaluations offer little general guidance, and a new approach to evaluating and comparing community detection algorithms is needed.

Here, we present a broad and comprehensive comparison of the performance of 16 state-of-the-art community detection methods and we evaluate the degree to and circumstances under which they under- or over-fit to network data. We evaluate these methods using a novel corpus of 572 real-world networks from many scientific domains, which constitutes a realistic and structurally diverse benchmark for evaluating and comparing the practical performance of algorithms. We characterize  each algorithm's performance (i)~relative to general theoretical constraints, (ii)~on the practical task of link prediction (a kind of cross-validation for network data), and (iii)~on a new task we call link description. The tradeoff between these two tasks is analogous to the classic bias-variance tradeoff in statistics and machine learning, adapted to a network setting in which pairwise interactions violate independence assumptions.

In both link description and link prediction, some fraction of a network's observed edges are removed before communities are detected, much like dividing a data set into training and test sets for cross validation. We then score how well the identified communities (and any corresponding model parameters the method returns) predict the existence of either the remaining observed edges (link description) or the removed edges (link prediction). By design, no algorithm can be perfectly accurate at both tasks, and the relative performance becomes diagnostic of a method's tendency to overfit or underfit to data.
Hence, as in a non-relational learning setting, a method can be said to overfit to the data if its accuracy is high on the training data but low for the test data, and it can be said to underfit if its accuracy is low on both training and test data. 

Our results show that (i)~algorithms vary widely both in the number of communities they find and in their corresponding composition, given the same input, (ii)~algorithms can be clustered into distinct high-level groups based on similarities of their outputs on real-world networks, (iii)~algorithmic differences induce wide variation in accuracy on link-based learning tasks, and (iv)~no algorithm is always the best at such tasks across all inputs.
Finally, we introduce and apply a diagnostic that uses the performance on link prediction and link description to evaluate a method's general tendency to under- or over-fitting in practice.

Our results demonstrate that many methods make uncontrolled tradeoffs that lead to overfitting on real data.  Across methods, Bayesian and regularized likelihood methods based on SBM tend to perform best, and a minimum description length (MDL) approach to regularization~\cite{peixoto2013parsimonious} provides the best general learning algorithm. On some real-world networks and in specific settings, other approaches perform better, which illustrates the NFL's relevance for community detection in practice. That is, although the SBM with MDL regularization may be a good general algorithm for community detection, specialized algorithms can perform better when applied to their preferred inputs.

\section{Methods and Materials}
\label{sec:MM}

Despite well-regarded survey articles~\cite{porter:etal:2010review,fortunato2010review,peixoto2017bayesian} there are relatively few comparative studies for model selection techniques in community detection~\cite{zhang2012comparative,hoff2008modeling,hric:darst:fortunato:2014,le2015estimating,kawamoto2016comparative,dabbs2016comparison} and most of these consider only a small number of methods using synthetic data (also called ``planted partitions'') or select only a small number of real-world networks, e.g., the Zachary karate club network, a network of political blogs, a dolphin social network, or the NCAA 2000 schedule network. The narrow scope of such comparative studies has been due in part to the non-trivial nature both of implementing or obtaining working code for competing methods, and of applying them to a large and representative sample of real-world network data sets.

For example, Ref.~\cite{le2015estimating} compared several spectral approaches using planted partition networks and five small well-studied real-world networks. In contrast, Ref.~\cite{kawamoto2016comparative} carried out a broader comparative analysis of the number of clusters found by six different algorithms. 
The authors also introduce a generalized message passing approach for modularity maximization~\cite{zhang2014scalable} in which they either use modularity values directly or use leave-one-out cross-validation~\cite{kawamoto2016cross} to infer the number of clusters. These methods were only evaluated on planted partition models and a small number of real-world networks.
Ref.~\cite{dabbs2016comparison} proposed a multi-fold cross-validation technique similar to Refs.~\cite{chen2014network,hoff2008modeling} and compared results with other cross-validation techniques using synthetic data. Recently, Ref.~\cite{valles-catala_consistency_2017} showed that model selection techniques based on cross-validation are not always consistent with the most parsimonious model and in some cases can lead to overfitting. None of these studies compares 
methods on a realistically diverse set of networks, or provides general guidance on evaluating over- and under-fitting outcomes in community detection.

In general, community detection algorithms can be categorized into two general settings. The first group encompasses probabilistic models, which use the principled method of statistical inference to find communities. Many of these are variants on the popular stochastic block model (SBM).
Under this probabilistic generative model for a graph $G = (V, E)$ with the size $N:=|V|$, a latent variable denoting the node's community label $g_i\in\{1,...,k\}$, with prior distribution $q_a$ ($a \in \{1,...,k\}$), is assigned to each node $i \in V$. Each pair of nodes $i,j \in V \times V$ is connected independently with probability $p_{g_i , g_j}$. In the sparse case, where $M := |E| = O(N)$, the resulting network is locally tree-like and the number of edges between groups is Poisson distributed. However, a Poisson degree distribution does not match the heavy-tailed pattern observed in most real-world networks, and hence the standard SBM tends to find partitions that correlate with node degrees.
The degree-corrected stochastic block model (DC-SBM)~\cite{karrer2011stochastic} corrects this behavior by introducing a parameter $\theta_i$ for each node $i$ and an identifiability constraint on $\theta$. In this model, each edge $(i,j)$ exists independently with probability $p_{g_i , g_j}\theta_i \theta_j$. The aforementioned planted partition model for synthetic networks can simply be a special case of the SBM with $k$ communities, when $p_{g_i,g_j}=p_{\iin}$ if $g_i=g_j$ and $p_{\out}$ if $g_i \neq g_j$~\cite{condon2001algorithms}.

This first group of methods includes a variety of regularization approaches for choosing the number of communities, e.g., those based on penalized likelihood scores~\cite{daudin2008mixture,come2015model}, various Bayesian techniques including marginalization~\cite{newman2016estimating,yan2016bayesian}, cross-validation methods with probabilistic models~\cite{kawamoto2016cross,kawamoto2016comparative}, compression approaches like MDL~\cite{peixoto2013parsimonious}, and explicit model comparison such as likelihood ratio tests (LRT)~\cite{wang2015likelihood}.

The second group of algorithms encompasses non-probabilistic score functions. This group is more varied, and contains methods such as modularity maximization and its variants~\cite{newman2004finding,zhang2014scalable}, which maximizes the difference between the observed number of edges within groups and the number expected under a random graph with the same degree sequence; the map equation (Infomap)~\cite{rosvall2008maps}, which uses a two-level compression of the trajectories of random walkers to identify groups; and, various spectral techniques~\cite{Krzakala2013,le2015estimating}, which seek a low-rank approximation of a noisy but roughly block-structured adjacency matrix, among others.

Methods in both groups can differ by whether the number of communities $k$ is chosen explicitly, as a parameter, or implicitly, either by an assumption embedded within the algorithm or by a method of model complexity control. In fact, the distinction between explicit and implicit choices can be subtle
as evidenced by a recently discovered equivalence between one form of modularity maximization (traditionally viewed as choosing $k$ implicitly) and one type of SBM (which typically makes an explicit choice)~\cite{newman2016community}. For the interested reader, a brief survey of model selection techniques for community detection is presented in Appendix~\ref{secA:MSA}.

\begin{table}[t!]\addtolength{\tabcolsep}{-5pt}
\caption{Abbreviations and descriptions of 16 community detection methods.}
\vspace*{-3mm}
\centering
 \begin{tabular}{p{2.42cm} p{0.90cm} p{5.4cm}} 
 
 \hline
Abbreviation & Ref. &  Description  \\ [0.5ex] 
 \hline\hline
Q&\cite{newman2004finding} & Modularity, Newman-Girvan  \\  \hline
Q-MR&\cite{newman2016community} & Modularity, Newman's multiresolution    \\  \hline
Q-MP & \cite{zhang2014scalable}& Modularity, message passing   \\  \hline
Q-GMP &\cite{kawamoto2016comparative} & Modularity, generalized message passing with CV-LOO as model selection    \\  \hline
B-NR (SBM)&\cite{newman2016estimating} & Bayesian, Newman and Reinert   \\  \hline
B-NR (DC-SBM)&\cite{newman2016estimating} & Bayesian, Newman and Reinert   \\  \hline
B-HKK (SBM) &\cite{hayashi2016tractable} & Bayesian, Hayashi, Konishi and Kawamoto \\  \hline
cICL-HKK (SBM)&\cite{hayashi2016tractable}& Corrected integrated classification likelihood \\  \hline
Infomap&\cite{rosvall2008maps} & Map equation\\  \hline
MDL (SBM)&\cite{peixoto2013parsimonious}  & Minimum description length \\  \hline
MDL (DC-SBM)&\cite{peixoto2013parsimonious}  & Minimum description length \\  \hline
S-NB&\cite{Krzakala2013} & Spectral with non-backtracking matrix  \\  \hline
S-cBHm&\cite{le2015estimating} & Spectral with Bethe Hessian, version m  \\  \hline
S-cBHa&\cite{le2015estimating} & Spectral with Bethe Hessian, version a  \\  \hline
AMOS&\cite{chen2016phase} &  Statistical test using spectral clustering   \\  \hline
LRT-WB (DC-SBM)&\cite{wang2015likelihood} & Likelihood ratio test\\  \hline
 \hline
 \label{table:abbr}
 \end{tabular}
\end{table}

In this study, a central aim is to develop and apply a principled statistical method to evaluate and compare the degree to which different community detection algorithms under- or over-fit to data. Toward this end, we compare the results of a large number of algorithms in order to illustrate the different kinds of behaviors that emerge, and to ensure that our results have good generality. For this evaluation,
we selected a set of 16 representative and state-of-the-art approaches that spans both general groups of algorithms (see Table~\ref{table:abbr}). This set of algorithms is substantially larger and more methodologically diverse than any previous comparative study of community detection methods and covers a broad variety of approaches. To be included, an algorithm must have had reasonably good computational complexity, generally good performance, and an available software implementation. Because no complete list of community detection algorithms and available implementations exists, candidate algorithms were identified manually from the literature, with an emphasis on methodological diversity in addition to the above criteria.

From information theoretic approaches we selected MDL~\cite{peixoto2013parsimonious} and Infomap~\cite{rosvall2008maps}. From the regularized likelihood approaches, we selected the corrected integrated classification likelihood (cICL-HKK)~\cite{hayashi2016tractable}.
From among the Bayesian methods Newman and Reinert's Bayesian (\mbox{B-NR})~\cite{newman2016estimating} and Hayashi, Konishi and Kawamoto's Bayesian (B-HKK)~\cite{hayashi2016tractable} are selected. From among the modularity based approaches, we selected Newman's multiresolution modularity (\mbox{Q-MR})
~\cite{newman2016community}, the classic Newman-Girvan modularity (Q)~\cite{newman2004finding} (both fitted using the Louvain method~\cite{blondel2008fast}), the Zhang-Moore message passing modularity (\mbox{Q-MP})~\cite{zhang2014scalable}, and the Kawamoto-Kabashima generalized message passing modularity algorithm (\mbox{Q-GMP})~\cite{kawamoto2016comparative}. (We consider a cross-validation technique called leave-one-out (CV-LOO)~\cite{kawamoto2016cross} to be the model selection criterion of \mbox{Q-GMP}.) From the spectral methods, we selected the non-backtracking (\mbox{S-NB})~\cite{Krzakala2013} and Bethe Hessian~\cite{saade2014spectral,le2015estimating} approaches. For the latter method, we include two versions, \mbox{S-cBHa} and \mbox{S-cBHm}~\cite{le2015estimating}, which are corrected versions of the method proposed in Ref.~\cite{saade2014spectral}.  
From among the more traditional statistical methods, we selected AMOS~\cite{chen2016phase} and a likelihood ratio test (\mbox{LRT-WB})~\cite{wang2015likelihood}. 

Examples of algorithms that were not selected under the criteria listed above include the infinite relational model (IRM)~\cite{kemp2006learning}, a nonparametric Bayesian extension of SBM, along with a variant designed to specifically detect assortative communities~\cite{morup2012bayesian}, the order statistics local optimization method (OSLOM)~\cite{lancichinetti2011finding}, which finds clusters via a local optimization of a significance score, and a method based on semidefinite programming~\cite{montanari2016semidefinite}.
%
Two related classes of algorithms that we do not consider are those that return either hierarchical~\cite{salespardo:etal:2007,clauset:moore:newman:2008,rosvall2011multilevel,peixoto2014hierarchical} or mixed-membership communities~\cite{airoldi2008mixed}. Instead, we focus on traditional community detection algorithms, which take a simple graph as input and return a ``hard'' partitioning of the vertices. As a result, hierarchical decompositions or mixed membership outputs are not directly comparable, without additional assumptions.

As a technical comment, we note that the particular outputs of some algorithms depend on the choice of a prior distribution, as in the Bayesian approaches, or on some details of the implementation. For example, the MDL and Bayesian integrated likelihood methods are mathematically equivalent for the same prior~\cite{peixoto2017bayesian}, but can produce different outputs with 
different priors and implementations (see below). However, the qualitative results of our evaluations are not affected by these differences. Finally, we note that the link prediction task is carried out using the learned models themselves, rather than using sampling methods, which improves comparability despite such differences.

\begin{figure}[t!]
\centering
\begin{tabular}{cc}
\includegraphics[width=1\columnwidth]{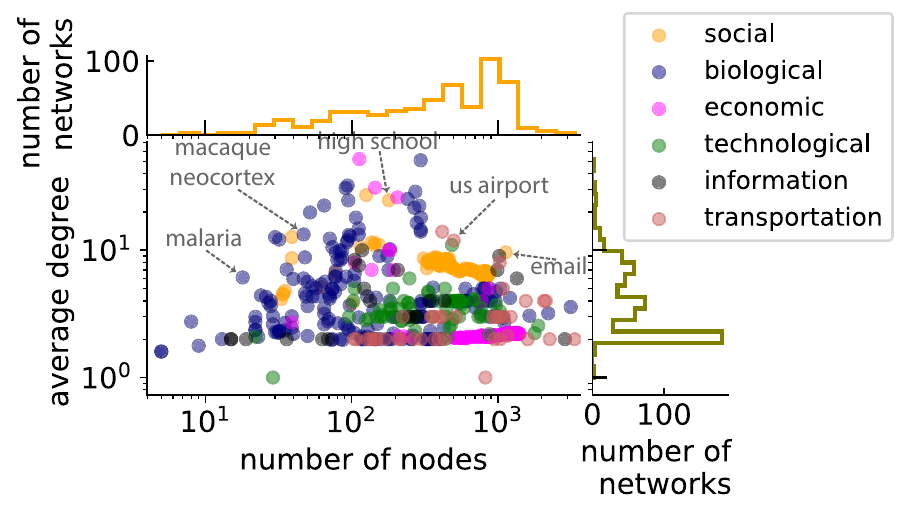}
\end{tabular}
\vspace{-0.4cm}
\caption{Average degree versus number of nodes for the corpus of 572 real-world networks studied here. Networks were drawn from the Index of Complex Networks (ICON)~\cite{clauset2016ICON}, and include social, biological, economic, technological, information, and transportation graphs.}
\label{fig:AD}
\end{figure}
 
\begin{figure*}[h]
     \begin{center}
            \includegraphics[width=1\textwidth]{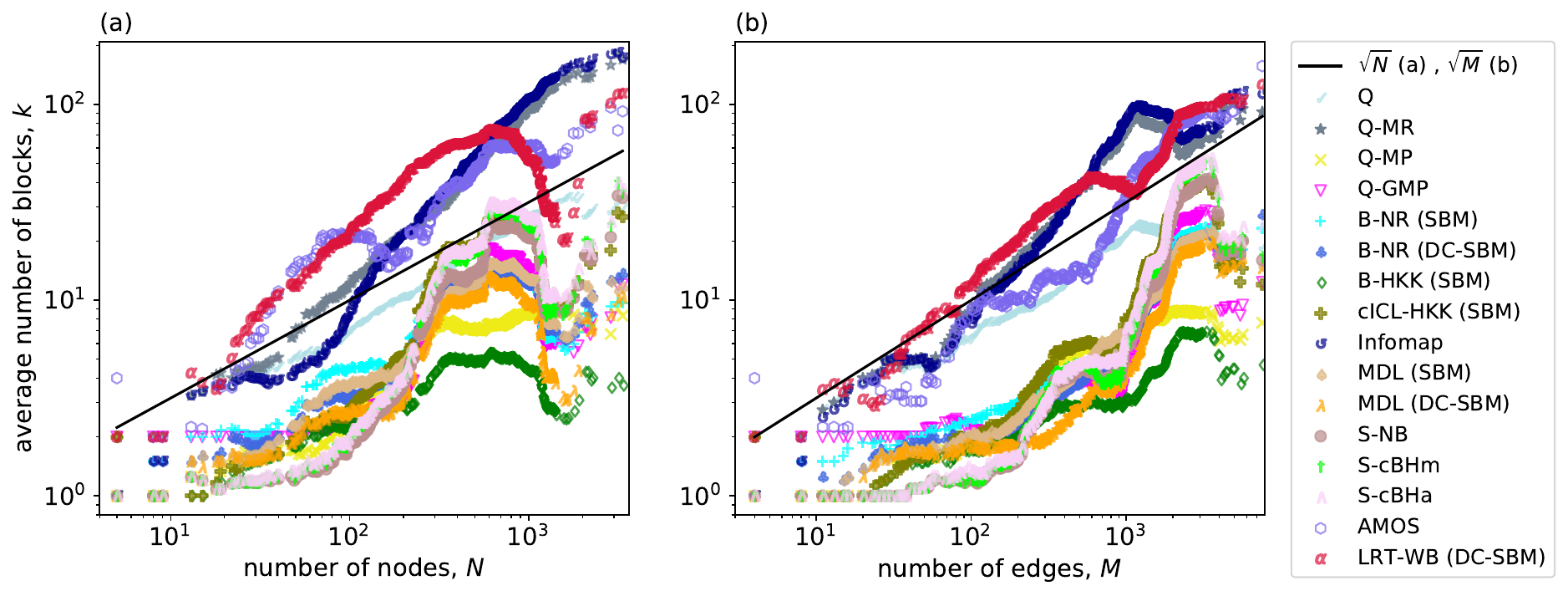}
    \end{center}
    \vspace*{-5mm}
    \caption{The average number of inferred communities, for 16 state-of-the-art methods (see Table~\ref{table:abbr}) applied to 572 real-world networks from diverse domains, versus the (a) number of nodes $N$, with a theoretical prediction of $\sqrt{N}$, or (b) number of edges $M$, with a theoretical prediction of $\sqrt{M}$.}%
   \label{fig:AD_NE}
\end{figure*}

To evaluate and compare the behavior of these community detection algorithms in a practical setting, we introduce and exploit the ``CommunityFitNet corpus,'' a novel data set\footnote{Available at \url{https://github.com/AGhasemian/CommunityFitNet}} containing 572 real-world networks drawn from the Index of Complex Networks (ICON)~\cite{clauset2016ICON}. The CommunityFitNet corpus spans a variety of network sizes and structures, with 22\% social, 21\% economic, 34\% biological, 12\% technological, 4\% information, and 7\% transportation graphs (Fig.~\ref{fig:AD}). Within it, the mean degree of a network is roughly independent of network size, making this a corpus of sparse graphs. In our analysis, we treat each graph as being simple, meaning we ignore the edge weights and directions. If a graph is not connected, we consider only its largest component. 

\section{Number of Communities in Theory and Practice}
\label{sec:NCTP}
\subsection{In Theory}
\label{sec:IT}
A key factor in whether some community detection method is over- or under-fitting to a network is its selection of the number of clusters or communities $k$ for the network. In the community detection literature, most of the consistency theorems, which provide guarantees on the fraction of mislabeled nodes, apply to dense networks only. For example, Ref.~\cite{choi2012stochastic}, proposes that the fraction of misclassified nodes converges in probability to zero under maximum likelihood fitting, when the number of clusters grows no faster than $O(\sqrt{N})$ and when the average degree grows at least poly-logarithmically in $N$.

However, most real-world networks are sparse~\cite{leskovec2008statistical,jacobs2015assembling}, including all networks in the CommunityFitNet corpus (Fig.~\ref{fig:AD}), meaning results for dense networks are inapplicable. For sparse networks, several lines of mathematical study argue that the maximum number of detectable clusters is $O(\sqrt{N})$, explicitly~\cite{peixoto2013parsimonious,peixoto2017nonparametric,chen2016statistical} or implicitly as an assumption in a consistency theorem~\cite{ames2014guaranteed,chen2014improved,alon1998finding}.

For example, in the planted $k$-partition, the expected number of recoverable clusters grows like $O({\sqrt{N}})$~\cite{ames2014guaranteed,chen2014improved}. (For convex optimization on the planted $k$-partition model, a tighter $O(\log N)$ bound on the number of clusters has also been claimed~\cite{ailon2015iterative}, although this result is not rigorous.) This theoretical limit is remarkably similar to the well-known resolution limit in modularity~\cite{fortunato2007resolution}, which shows that modularity maximization will fail to find communities with sizes smaller than $O(\sqrt{M})$. Hence, the expected number of modularity communities in a sparse graph is also $O(\sqrt{M})$.
An argument from compression leads to a similar bound on $k$~\cite{peixoto2013parsimonious,peixoto2017nonparametric}. Specifically, the model complexity of a stochastic block model is $\Theta(k^2)$, which under a minimum description length analysis predicts that $k=\Theta(\sqrt{M})$. This statement can also be generalized to regularized likelihood and Bayesian methods. Although none of these analyses is fully rigorous, they do point to a similar theoretical prediction:\ the number of recoverable communities in a real sparse network should grow like $O(\sqrt{N})$. Different algorithms, of course,  may have different constants of proportionality or different sub-asymptotic behaviors. In our evaluations, we use a constant of 1 as a common reference point.

As a technical comment, we note that the maximum number of clusters found is not the same as the number of identifiable clusters under the information-theoretic limit to detectability~\cite{decelle2011asymptotic}. For example, as a result of a resolution limit, an algorithm might merge two clusters, but still infer the remaining clusters correctly. In other words, the network's communities exist in the detectable regime but the output has lost some resolution.

\subsection{In Practice}

\label{sec:IP}
We applied our set of 16 community detection methods to the 572 real-world networks in the CommunityFitNet corpus. For methods with free parameters, values were set as suggested in their source papers. We then binned networks by their size $N$ or $M$ and for each method plotted the average (Fig.~\ref{fig:AD_NE}) and maximum number (Fig.~\ref{fig:MD_NE}; see Appendix) of inferred communities as a function of the number of nodes $N$ and edges $M$.

In both figures, solid lines show the theoretically predicted trends of $\sqrt{N}$ and $\sqrt{M}$. Two immediate conclusions are that (i) the actual behavior of different algorithms on the same input is highly variable, often growing non-monotonically with network size and with some methods finding 10 times as many communities as others, but (ii) overall, the number of communities found does seem to grow with the number of edges, and perhaps even roughly like the $\sqrt{M}$ pattern predicted by different theoretical analyses (Section~\ref{sec:IT}). Furthermore, the empirical trends are somewhat more clean when we consider the number of communities $k$ versus the number of edges $M$ (Fig.~\ref{fig:AD_NE}b and Fig.~\ref{fig:MD_NE}b; see Appendix), suggesting that the mean degree of a network impacts the behavior of most algorithms.

From this perspective, algorithms visually cluster into two groups, one of which finds roughly 2-3 times as many communities as the other. The former group contains the methods of Q, Q-MR, Infomap, LRT-WB and AMOS, all of which find substantially more clusters than methods in the latter group, which includes B-NR, B-HKK, cICL-HKK, MDL, spectral methods, Q-MP and Q-GMP. In fact, first group of methods often return many more clusters than we would expect theoretically, which suggests the possibility of consistent overfitting. As a small aside, we note that the expected number of communities found by Q-MR and Q are different, because they are known to have different resolution limits~\cite{kumpula2007limited}. More generally, the aforementioned groups, and their tendency to find greater or fewer communities aligns with our dichotomy of non-probabilistic (more communities) versus probabilistic (fewer communities) approaches. Additionally, we note that the AMOS method failed to converge on just over half of the networks in the CommunityFitNet corpus, returning an arbitrary maximum value instead of $k$. Because of this behavior, we excluded AMOS from all subsequent analysis.

In Ref.~\cite{kawamoto2016comparative}, the authors show empirically that Infomap and the Louvain method for modularity maximization tends to overfit the number of clusters planted in synthetic networks with weak community structure. The results shown here on our large corpus of real-world networks are consistent with these previous results, indicating that both modularity and Infomap tend to find substantially more communities compared to other methods. Figure~\ref{fig:MD_NE} (see Appendix), shows more clearly that the maximum number of clusters detected by Q-MR and Infomap are nearly identical. Furthermore, these methods find the same average number of clusters over 572 networks (Fig.~\ref{fig:AD_NE}). This behavior has not previously been noted, and suggests that \mbox{Q-MR} and Infomap may have the same or very similar resolution limits.

In general, algorithms with similar formulations or that are based on similar approaches show similar behavior in how $k$ varies with $M$. For instance, beyond \mbox{Q-MR} and Infomap's similarity, we also find that MDL, various regularized likelihood methods, and the Bayesian approaches find similar numbers of communities. Spectral methods appear to behave similarly, on average (Fig.~\ref{fig:AD_NE}), to the Bayesian approaches. However, spectral approaches do seem to overfit for large network sizes, by finding a maximum number of communities that exceeds theoretical predictions, in contrast to the Bayesian approaches.

Looking more closely at similar methods, we observe small differences in the number of clusters $k$ they return as a function of network size $N$, which must be related to differences in their implicit assumptions. For example \mbox{B-HKK}, \mbox{B-NR} and \mbox{cICL-HKK} often agree on the number of communities for networks with a smaller number of edges, but they disagree for networks with a larger number of edges (Fig.~\ref{fig:AD_NE}). Due to a more exact Laplace approximation with higher order terms, \mbox{B-HKK} penalizes the number of clusters more than \mbox{B-NR} and \mbox{cICL-HKK}, which limits the model space of \mbox{B-HKK} to smaller models that correspond to fewer communities. This tradeoff is a natural one, as approaches that penalize a model for greater complexity, like in \mbox{B-HKK}, are intended to reduce the likelihood of overfitting, which can in turn increase the likelihood of underfitting.

Returning to the \mbox{Q-MR} method, we inspect its results more carefully to gain some insight into whether it is overfitting or not. Ref.~\cite{newman2016community} proves that \mbox{Q-MR} is mathematically equivalent to a \mbox{DC-SBM} method on a $k$-planted partition space of models. The \mbox{Q-MR} algorithm works implicitly like a likelihood-maximization algorithm, except that it chooses its resolution parameter, which sets $k$, by iterating between the Q and \mbox{DC-SBM} formulations of the model. Evidently, this approach does not limit model complexity as much as a regularized likelihood and tends to settle on a resolution parameter that produces a very large number of communities. This behavior highlights the difficulty of characterizing the underlying tradeoffs that drive over- or under-fitting in non-probabilistic methods. We leave a thorough exploration of such questions for future work.

The variation across these 16 methods of the average and maximum number of communities found, provides suggestive evidence that some methods are more prone to over- or under-fitting than others, in practice. The broad variability of detected communities by different methods applied to the same input is troubling, as there is no accepted procedure for deciding which output is more or less useful.

\section{Quantifying algorithm similarity}

\begin{figure*}[t!]
\centering
\begin{tabular}{c}
\includegraphics[width=1.5\columnwidth]{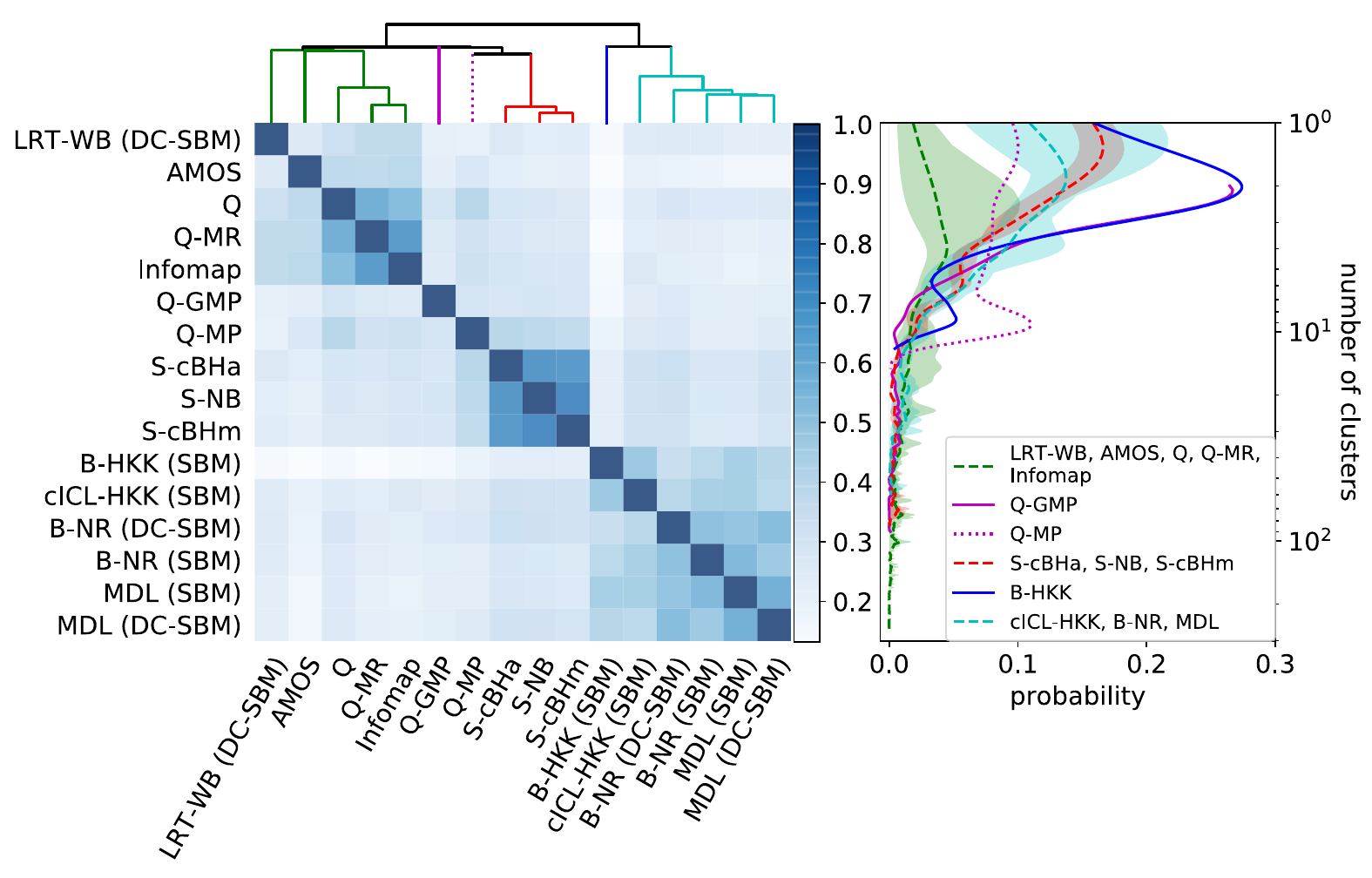}
\end{tabular}
\caption{ A clustering of community detection algorithms into distinct high-level groups based on the similarities of their outputs on real-world networks. (a) The mean adjusted mutual information (AMI) between each pair of methods for communities they recovered on each network in the CommunityFitNet corpus. Rows and columns have been ordered according to the results of a hierarchical clustering of the AMI matrix, after applying a Gaussian kernel with parameter $\sigma^2=0.3$. (b) Density plots showing the distribution of the number of inferred communities $k$ for groups of similar algorithms.}
\label{fig:dend-NMI}
\end{figure*}

Although algorithms can be divided into groups based on their general approach, e.g., probabilistic and non-probabilistic methods (see Section~\ref{sec:intro}), such conceptual divisions may not reflect practical differences when applied to real data. Instead, a data-driven clustering of algorithms can be obtained by comparing the inferred labels of different methods applied to a large and consistent set of real-world networks. That is, we can use our structurally diverse corpus to empirically identify groups of algorithms that produce similar kinds of community structure across different data sets. (We note that this kind of comparative analysis has previously been employed to better characterize the behavior of different time series methods~\cite{fulcher2013highly}.) We quantify algorithm similarity by computing the mean adjusted mutual information (AMI)~\cite{vinh2010information} between each pair of methods for the communities they recover on each network in the CommunityFitNet corpus. We then apply a standard hierarchical clustering algorithm to the resulting matrix of pairwise similarities in algorithm output (Fig.~\ref{fig:dend-NMI}a). Using the unadjusted or normalized mutual information (NMI) yields precisely the same clustering results, indicating that these results are not driven by differences in the sizes of the inferred communities, which are broadly distributed (Fig.~\ref{fig:dend-NMI}b).

The derived clustering of algorithms shows that there is, indeed, a major practical difference in the composition of communities found by probabilistic versus non-probabilistic methods. In fact, methods based on probabilistic models typically find communities that are more similar to those produced by other probabilistic methods, than to those of any non-probabilistic method, and vice versa. This high-level dichotomy indicates a fairly strong division in the underlying assumptions of these two classes of algorithms.

The non-probabilistic methods group further subdivides into subgroups of spectral algorithms (\mbox{S-cBHa}, \mbox{S-NB}, and \mbox{S-cBHm}), consensus-based modularity algorithms (\mbox{Q-MP} and \mbox{Q-GMP}), traditional statistical methods (AMOS and \mbox{LRT-WB}), and finally other non-probabilistic methods (including Infomap, Q, and \mbox{Q-MR}). The fact that algorithms themselves cluster together in the kind of outputs they produce has substantial practical implications for any application that depends on the particular composition of a network clustering. It also highlights the subtle impact that different classes of underlying assumptions can ultimately have on the behavior of these algorithms when applied to real-world data.

\section{Evaluating Community Structure Quality}
\label{sec:ECSQ} 

To evaluate and compare the quality of the inferred clusters for a particular network, we need a task that depends only on a network's connectivity and that can reveal when a method is over- or under-fitting these data. (Recall that the NFL and ``no ground truth'' theorems of Ref.~\cite{peel2017ground} imply that a comparison based on node metadata cannot be reliable.) For relational data, a common approach uses a kind of network cross-validation technique, called link prediction~\cite{lu2011link}, in which some fraction of the observed edges in a network are ``held out" during the model-fitting stage, and their likelihood estimated under the fitted model.

We note, however, that there is as yet neither a consensus about how to design such a task optimally nor a theoretical understanding of its relationship to model fit. For example, it was recently shown that selecting the most parsimonious probabilistic model in community detection, by maximizing model posterior probability, can correlate with selecting the model with highest link prediction accuracy~\cite{valles-catala_consistency_2017}. However, these same results show that it is possible to construct networks, i.e., adversarially, in which the most plausible model (in the sense of posterior probability) is not the most predictive one, and hence improving predictive performance can also lead to overfitting.
Furthermore, the theoretical implications are unknown for distinct approaches to construct a held-out data set from a single network, for example, holding out a uniformly random subset of edges or all edges attached to a uniformly random subset of nodes. Although there are strong theoretical results for cross-validation and model selection for non-relational data, whether these results extend to networks is unclear as relational data may violate standard independence assumptions. Theoretical progress on this subject would be a valuable direction of future research.

Here, we introduce an evaluation scheme based on a pairing of two complementary network learning tasks: a link prediction task, described above and in Box~\hyperref[table:LP]{1}, and a new task we call link description, described below and in Box~\hyperref[table:LD]{2}. The goal of these tasks is to characterize the behavior of methods in general, i.e., a method's general tendency to over- or under-fit across many real networks, rather than to evaluate the quality of a fit to any particular network. A key feature of this scheme is that a method cannot be perfect at both tasks, and each method's tradeoff in performance across them creates a diagnostic to evaluate the method's
tendency to over- or under-fit real data.


In our evaluation, each method uses a score function to estimate the likelihood that a particular pair of nodes $ij$ should be connected. Most algorithms optimize some underlying objective function in order to sort among different community partitions. In our main evaluation, we use model-specific score functions, which are based on the method's own objective function. This choice ensures that each method makes estimates that are aligned with its underlying assumptions about what makes good communities. We also compare these model-specific results with those derived from a SBM-based scoring function in Appendix~\ref{secA:MA-LPLD}. This comparison to a fixed reference point allows us to better distinguish between poor generalizability being caused by a low-quality score function
and the selection of a low-quality partition or set of communities.

\subsection{Model-specific Link Prediction and Description}
\label{sec:MS-LPLD}
\subsubsection*{Link prediction} 
\label{sec:MS-LP}

When a graph $G=(V,E)$ has been sampled to produce some $G'=(V,E')$, where $E' \subset E$, the goal of link prediction is to accurately distinguish missing links (true positives) from non-edges (true negatives)  within the set of unobserved connections $ij \in V\times V \smallsetminus E'$.
Link prediction is thus a binary classification task and its general accuracy can be quantified by the area under the ROC curve (AUC).

For our evaluation, we parameterize this classification accuracy by $\alpha\in(0,1)$, which determines the fraction of edges ``observed'' (equivalently, the density of sampled edges) in $G'=(V,E')$, where $|E'|=\alpha |E|$ is a uniformly random subset of edges in the original graph $G$. For a given method $f$, its AUC as a function of $\alpha$, which we call an ``accuracy curve,'' shows how $f$ performs across a wide variety of such sampled graphs, ranging from when very few edges are observed ($\alpha\to0$) to when only a few edges are missing ($\alpha\to1$).

Each network $G$ in our corpus produces one such accuracy curve, and we obtain a single ``benchmark performance curve'' for each method by computing the mean AUC at each value of $\alpha$, across curves produced by the networks in the CommunityFitNet corpus. When computing the AUC, we break ties in the scoring function uniformly at random. Box~\hyperref[table:LP]{1} describes the link prediction task in detail.

In this setting, the AUC is preferred over precision-recall because we are interested in the general performance of these classifiers, rather than their performance under any specific prediction setting. Evaluating other measures of accuracy is beyond the scope of this study. Comparing benchmark performance curves across community detection methods quantifies their relative generalizability and predictiveness, and allows us to assess the quality of choice each method makes for the number of clusters it found in Section~\ref{sec:IP}.

\begin{tcolorbox}[breakable,label=table:LP,float = t!,size=title,
  colback=blue!2!white,colframe=blue!5!black,fonttitle=\bfseries,
  title=Box 1: Link Prediction Benchmark,pad at break=1mm, break at=-\baselineskip/0pt]
  \begin{itemize}
  \item Let $\mathcal{G}$ be a corpus of networks, each defined as a graph $G=(V,E)$.
  \item For each $G\in\mathcal{G}$, define a ``sampled graph'' as \mbox{$G'=(V,E')$}, where $E' \!\subset\! E$, 
and \mbox{$|E'|=\alpha |E|$} for \mbox{$\alpha \in (0,1)$} is a uniformly random edge subset.
 \item Let $f$ denote a community detection method. 	
  \item Let $s_{ij}\in \mathbb{R}$ denote a score function specific to $f$ that assigns a numerical value to each potential missing link $ij \in V\times V \smallsetminus E'$, with ties broken uniformly at random.
  \item Define the accuracy of $f$, applied to $G'$ and measured by $s$, as the AUC on distinguishing missing links (true positives) $ij \in E \smallsetminus E'$ from non-edges (true negatives) $ij\in V\times V \smallsetminus E$.
  \item Define the link prediction ``accuracy curve'' to be the AUC of $f$ on a set of $G'$, for $0<\alpha<1$.
  \item Define the link prediction ``benchmark performance curve'' of $f$ to be the mean AUC of $f$ over all $G'\in\mathcal{G}$ at each $\alpha$.
  \end{itemize}
\end{tcolorbox}

In our link prediction and link description evaluations, 
we exclude \mbox{S-cBHa} and \mbox{S-cBHm} because they produce very similar results to the \mbox{S-NB} method, \mbox{LRT-WB} because of its high computational complexity, and AMOS and \mbox{Q-GMP} because of convergence issues. All other methods are included.

We now define a model-specific score function for each method (see Appendix for more details). Each score function uses the corresponding method $f$ to define a model-specific function $s_{ij}$ that estimates the likelihood that a pair of nodes $ij$ should be connected. For probabilistic methods, the natural choice of score function is simply the posterior probability the model assigns to a pair $i,j$. For non-probabilistic methods, we constructed score functions that reflected the underlying assumptions of the corresponding method, without introducing many additional or uncontrolled assumptions.

For regularized likelihood/Bayesian approaches of the SBM (cICL-HKK, B-NR and B-HKK), we follow the same scoring rule as in Ref.~\cite{guimera2009missing}. Specifically, the score $s_{ij}$ assigned to an unobserved edge between nodes $i$ and $j$, for $(i,j) \notin E'$ with community assignments of $g_i$ and $g_j$, respectively, is given by $s_{ij} = \dfrac{\ell_{g_i,g_j}+1}{r_{g_i,g_j}+2}$, where $\ell_{g_i,g_j}$ is the number of edges in the observed network $G'$ between the groups $g_i$ and $g_j$ and $r_{g_i,g_j}$ is the maximum possible number of links between the groups $g_i$ and $g_j$. For DC-SBM we define $s_{ij} = \theta_i\theta_j\ell_{g_i,g_j}$, where $\theta_i$ is the normalized degree of node $i$ with respect to total degree of its type as the maximum likelihood estimation of this parameter.
For all the Q methods, Infomap, and MDL we define the scores as the contribution that the added unobserved edge would make to their corresponding partition score functions. For the Q methods, we compute the increase in the modularity score due to the added unobserved edge, while for Infomap and MDL we compute the decrease of their objective functions. For each of these methods, the contribution of each pair of nodes is computed under the partition obtained prior to adding the candidate pair as an edge.

There is no commonly accepted link prediction approach for spectral clustering algorithms that is independent of metadata. Although there are some non-linear embedding methods for link prediction like node2vec~\cite{grover2016node2vec}, here we focus on linear decomposition techniques. For spectral clustering, we introduce and use a new link prediction technique based on eigenvalue decomposition. Let the adjacency matrix of the observed graph $G'$ be denoted by $A'$. This matrix can be decomposed as $A'=V \Lambda V^T$, where $V=[v_1 v_2 \hdots v_N]$ with $v_i$ as $i$th eigenvector of matrix $A'$ and matrix $\Lambda=\text{diag}[\lambda_1,\lambda_2,\hdots,\lambda_N]$ is the diagonal matrix of eigenvalues of $A'$, where $\lambda_1\geq \lambda_2 \geq \hdots \geq \lambda_N$.

To define a new scoring function, we use a low-rank matrix approximation of $A'$ using the $k$ largest eigenvalues and their corresponding eigenvectors, i.e., we let $\hat{A'} = [v_1 v_2 \hdots v_k]\text{diag}[\lambda_1,\lambda_2,\hdots,\lambda_k][v_1 v_2 \hdots v_k]^T$, where $k$ can be inferred using a model selection spectral algorithm. The spectral method scoring rule $s_{ij}$ assigned to an unobserved edge between nodes $i$ and $j$, for $(i,j) \notin E'$, is the corresponding entry value in low-rank approximation. We note that alternative constructions exist for a spectral scoring function that meets the aforementioned criteria. 
	For instance, one could use the low-rank approximation via the non-backtracking matrix itself, but such a function would be quite non-trivial. On the other hand, the SBM-based and model-specific performance comparison given in Appendix~\ref{secA:MA-LPLD} indicates that the score function we use for this algorithm (\mbox{S-NB}) performs well and it is not itself the cause of this algorithm's observed poor performance. Exploring the broader space of scoring functions for spectral or other methods remains an interesting direction for future work.
 
The performance of each method is assessed by numerically computing its benchmark performance curve, using the 572 real-world networks in the CommunityFitNet corpus. Exactly calculating a single accuracy curve for a sparse graph $G$ takes $\Omega(N^2)$ time, which is prohibitive for large networks. However, each AUC in a curve may be accurately estimated using Monte Carlo, because the AUC is mathematically equivalent to the probability that a uniformly random true positive is assigned a higher score than a uniformly random true negative. In all of our experiments, an accuracy of $\pm 0.01$ is sufficient to distinguish performance curves, requiring 10,000 Monte Carlo samples.

Community detection methods that are prone to overfitting (underfitting) will tend to find more (fewer) communities in a network than is optimal. Hence, the induced partition of the adjacency matrix into within- and between-group blocks will over- (under-) or under- (under-) estimate the optimal block densities. This behavior will tend to produce lower AUC scores for the prediction of uniformly held-out pairs in the corresponding evaluation set. That is, a lower benchmark performance curve indicates a greater general tendency to over- or under-fit on real-world networks.

\subsubsection*{Link description}
\label{sec:MS-LD}
 
The goal of link description is to accurately distinguish observed edges $E'$ (true positives) and observed non-edges $V\times V \smallsetminus E'$ (true negatives) within the set of all pairs $ij\in V\times V$. That is, link description asks how well a method learns an observed network, and it is also a binary classification task.

As in link prediction, we parameterize the sampled graph $G'$ by the fraction $\alpha$ of observed edges from the original graph $G$. And, we use  the same scoring functions to evaluate an algorithm's accuracy at learning to distinguish edges from non-edges. Then, using the networks in the CommunityFitNet corpus, we obtain a benchmark performance curve for each method. Box~\hyperref[table:LD]{2} describes the link description task in detail.

\begin{tcolorbox}[breakable,label=table:LD,float=t!,size=title,
  colback=blue!2!white,colframe=blue!5!black,fonttitle=\bfseries,
  title=Box 2: Link Description Benchmark,pad at break=1mm, break at=-\baselineskip/0pt]

  \begin{itemize}
  \item Let $\mathcal{G}$ be a corpus of networks, each defined as a graph $G=(V,E)$.
  \item For each $G\in\mathcal{G}$, define a ``sampled graph'' as \mbox{$G'=(V,E')$}, where $E' \!\subset\! E$, and \mbox{$|E'|=\alpha |E|$} for \mbox{$\alpha \in (0,1)$} is a uniformly random edge subset.
 \item Let $f$ denote a community detection method.
  \item Let $s_{ij}\in \mathbb{R}$ denote a score function specific to $f$ that assigns a numerical value to each potential edge $ij \in V\times V$, with ties broken uniformly at random.
  \item Define the accuracy of $f$, applied to $G'$ and measured by $s$, as the AUC on distinguishing observed edges (true positives) $ij \in E'$ from observed non-edges (true negatives) $ij\in V\times V \smallsetminus E'$.
  \item Define the link description ``accuracy curve'' to be the AUC of $f$ on a set of $G'$, for $0<\alpha<1$.
  \item Define the link description ``benchmark performance curve'' of $f$ to be the mean AUC of $f$ over all $G'\in\mathcal{G}$ at each $\alpha$.
 
  \end{itemize}

\end{tcolorbox}

Crucially, an algorithm cannot perform perfectly at both the link prediction and the link description tasks. If an algorithm finds a very good partition for distinguishing between observed edges and observed non-edges (link description), this partition must assign low scores to all of the observed non-edges. This fact implies that the same partition cannot also be very good for distinguishing between non-edges and missing edges (link prediction), as it must assign low scores to both. The link prediction and description tasks thus force an algorithmic tradeoff, similar to a bias-variance tradeoff, and the joint behavior of a method across the two tasks provides a means to evaluate a tendency to over- or under-fit to real data.

\begin{figure*}[t!]
     \begin{center}
            \includegraphics[width=1\textwidth]{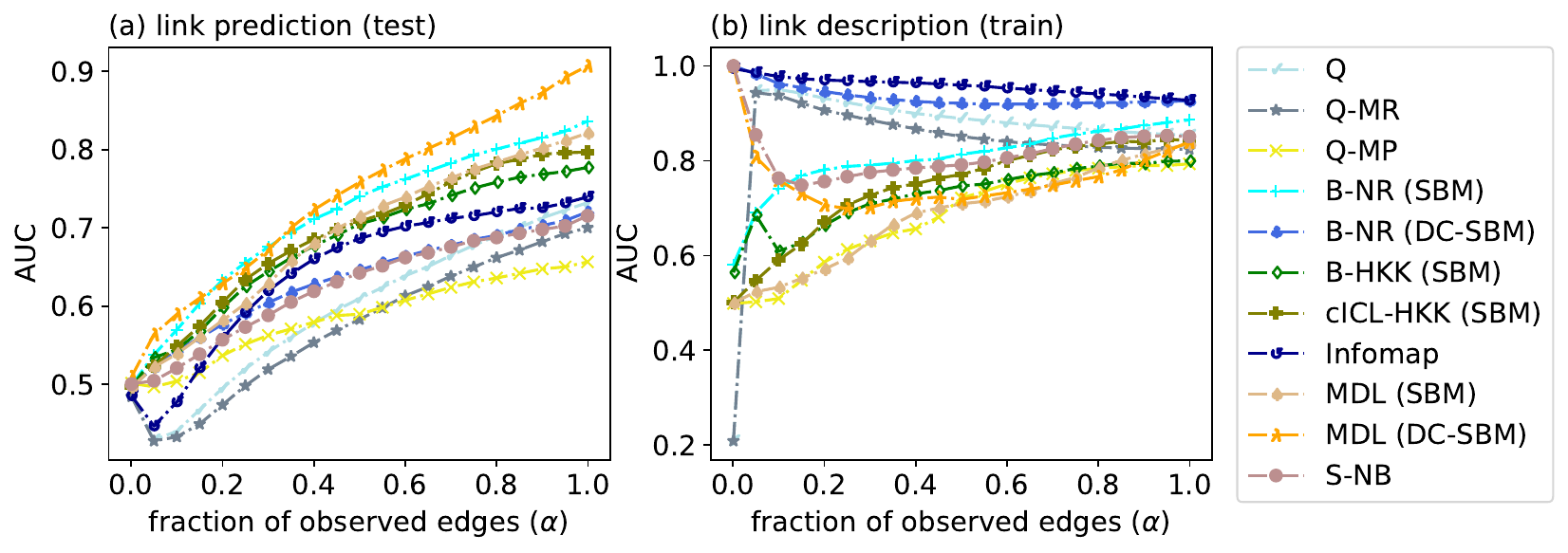}
    \end{center}
    \vspace*{-5mm}
    \caption{Benchmark performance curves using model-specific score functions for (a) link prediction and (b) link description tasks. Each curve shows the mean AUC for a different community detection method across 572 real-world networks for a given fraction $\alpha$ of observed edges in a network.}%
   \label{fig:LP}
\end{figure*}

\subsubsection*{Results}
 
The benchmark performance curves generally exhibit reasonable behavior:\ across the CommunityFitNet corpus networks, methods generally tend to improve their performance on both link prediction and link description tasks as the fraction of sampled edges varies from very low to very high (Fig.~\ref{fig:LP}), with some exceptions. For our purposes, we are specifically interested in the relative differences between the curves, which sheds new light on the degree to which methods tend to over- or under-fit on network data.

\medskip
In link prediction (Fig.~\ref{fig:LP}a), the low curves for Q methods reinforce our previous suggestion that these algorithms exhibit poor generalizability. Their performance is particularly poor when the edge density in the observed network is very low, e.g., when more than two-thirds of the edges are missing ($\alpha<0.3$) for \mbox{Q-MR} and Q. This behavior aligns well with the evidence in Section~\ref{sec:IP} that these methods tend to over-partition the network, finding many more communities than is optimal.

We also find that Q and \mbox{Q-MR} exhibit nearly identical benchmark performance curves for link prediction. Although Q has a larger resolution limit, which leads to fewer inferred clusters, our results suggest that Q still finds more communities than is optimal, especially compared to other methods. Evidently, these two methods tend to misinterpret noisy connectivity within communities as indicating the presence of substructure deserving of additional fitting, and they do this in a similar way, leading to similar numbers of communities and very similar link prediction curves. This behavior may reflect the common assumptions of these methods that communities are assortative (edges mainly within groups) and that the between-group edge densities are uniform. If this assumption is not reflected in the input network's actual large-scale structure, these methods can overfit the number of communities (Fig.~\ref{fig:AD_NE} and Fig.~\ref{fig:MD_NE}; see Appendix), subdividing larger groups to find a partition with the preferred structure.

The best benchmark performance curves for link prediction are produced by Bayesian methods (B-NR (SBM), B-HKK, and cICL-HKK) and MDL methods (both DC-SBM and SBM). And the SBM methods generally outperform the DC-SBM methods, except for the DC-SBM with MDL regularization, which yields the best overall benchmark curve for nearly every value of $\alpha$.

Such a difference is surprising since the number of inferred clusters for both Bayesian and regularized-likelihood methods is nearly identical (Fig.~\ref{fig:AD_NE}a,b), and the precise composition of the clusters is very similar (Fig.~\ref{fig:dend-NMI}). However, these methods use different score functions to estimate the likelihood of a missing edge, and evidently, those based on more general rules perform better at link prediction. For instance, B-NR (SBM) assigns the same scores to the links inside each cluster, whereas B-NR (DC-SBM) assigns higher scores to the links connected to high degree nodes. In \mbox{B-NR}, the emphasis on modeling node degrees by the DC-SBM score function leads to worse performance. In contrast, the MDL technique, while based on the same underlying probabilistic network model, assigns higher scores to edges that produce a better compression of the input data (shorter description length). Hence, the MDL score function prefers adjacencies that decrease the model entropy without increasing model complexity, meaning that it predicts missing links in places with better community structure. The MDL approach to controlling model complexity, particularly in the DC-SBM score function, is more restrictive than most Bayesian approaches, but it evidently leads to more accurate link prediction (Fig.~\ref{fig:LP}a).

The benchmark performance curves for Infomap, spectral clustering (S-NB), and B-NR (DC-SBM) are similar, especially for modest or larger values of $\alpha$, and are close to the middle of the range across algorithms. Furthermore, we find that the curves of B-HKK (SBM) and cICL-HKK (SBM) are similar to, but lower than B-NR (SBM). There are two possibilities for this behavior: ~(i) the number of inferred clusters is inaccurate, or (ii) these methods perform poorly at link prediction. Because the score functions of B-HKK (SBM), cICL-HKK (SBM), and \mbox{B-NR (SBM)} are similar, the lower link prediction benchmark performance is more likely caused by a low-quality set of inferred clusters, due to over- or under-fitting.

\smallskip
The benchmark results for link description (Fig.~\ref{fig:LP}b) show that B-HKK (SBM) and cICL-HKK (SBM) perform relatively poorly compared to most other methods, which suggests that they must tend to under-fit on networks. This behavior is likely driven by their larger penalty terms, e.g., compared to methods like B-NR (SBM), which will favor finding a smaller number of clusters in practice (Fig.~\ref{fig:AD_NE}). This behavior will tend to aid in link prediction at the expense of link description. We note that Ref.~\cite{hayashi2016tractable} introduced a better approximation for B-HKK (SBM)'s penalty terms, which might suggest that the method would find more optimal partitions in theory. However, our results show that this is not the case in practice, and instead this method illustrates a tradeoff in which a greater penalty for model complexity, by over-shrinking or over-smoothing the model space, can lead to poor performance in practical settings.

The best benchmark performance for link description is given by Infomap first, followed by the Bayesian technique B-NR (DC-SBM), and by modularity Q and \mbox{Q-MR}, all of which exhibit only middling performance curves for link prediction. This behavior suggests that all of these methods tend to overfit on networks. Others are better at link prediction compared to their relative performance at link description (MDL (DC-SBM), B-NR (SBM), cICL-HKK, and B-HKK), suggesting that these methods tend to well-fit on networks. And, some methods perform poorly on both tasks, such as \mbox{Q-MP}. These comparisons illustrate the intended tradeoff in our diagnostic between the link prediction and link description tasks, and provide a useful contrast for evaluating the practical implications of different kinds of underlying algorithmic assumptions.

The relative performance of Q, \mbox{Q-MR}, and Infomap versus other methods on these tasks provides an opportunity to understand how an algorithm's assumptions can drive over- and under-fitting in practice. By definition, the partitions found by Infomap and modularity-based methods like Q and \mbox{Q-MR} will tend to have highly assortative communities and a uniformly low density of edges between communities. Such a partition must perform poorly at modeling the few edges between these clusters; hence, as the density of these edges increases with $\alpha$, these methods' link description performance must tend to decrease.  In contrast, nearly all other methods generally perform better at link description as more edges are sampled, except for \mbox{B-NR} (DC-SBM), whose performance is relatively independent of $\alpha$. As a group, the probabilistic methods have the flexibility to model different rates of between-group connectivity, but this behavior requires sufficient observed connectivity in order to estimate the corresponding densities. As a result, these models are less data efficient than are modularity and spectral methods at describing the observed structure, especially in the sparsely-sampled regime (low $\alpha$).

The exception to this interpretation is \mbox{Q-MP}, which exhibits a poor performance on both tasks (Fig.~\ref{fig:LP}a,b). These tendencies can be understood in light of the relatively small number of communities \mbox{Q-MP} tends to find (Fig.~\ref{fig:AD_NE} and Fig.~\ref{fig:MD_NE}; see Appendix), which suggests that it tends to substantially under-fit to the data. In fact, \mbox{Q-MP} uses a consensus of many high-modularity partitions in order to explicitly avoid overfitting to spurious communities. Evidently, this strategy tends to find far fewer communities than is optimal for either link description or link prediction. The \mbox{Q-GMP} method may perform better, as it controls the bias-variance tradeoff through a learning phase on its parameters. Due to poor convergence behavior with this method, however, we leave this line of investigation for future work.

\begin{table*}[t!]\addtolength{\tabcolsep}{-5pt}
\caption{Summary of results for 16 algorithms (Table~\ref{table:abbr}) on the number of communities $k$ (Fig.~\ref{fig:AD_NE}b), the algorithm group the output is most similar to (Fig.~\ref{fig:dend-NMI}), benchmark performance on link prediction (Fig.~\ref{fig:LP}a) and link description (Fig.~\ref{fig:LP}b), and an overall assessment of its tendency to over- or under-fit.}
\vspace{-3mm}
\centering
 \begin{tabular}{p{2.6cm}p{2.3cm}p{2.5cm}p{2.5cm}p{2.5cm}p{2.5cm}} 
 \hline
& Number of & Partition & Link Prediction~~~ & Link Description~~~ &  \\ 
Algorithm & Communities, $k$ ~~~ & Type & Benchmark & Benchmark & Overall Fit \\ [0.5ex] 
 \hline\hline
Q & larger & non-probabilistic ~~~ & poor & good & over fits  \\  \hline
Q-MR & larger & non-probabilistic & poor & good & over fits    \\  \hline
Q-MP & smaller & \pbox{2.5cm}{spectral/ \\non-probabilistic}  & poor & poor & under fits   \\  \hline
Q-GMP & smaller &  \pbox{2.5cm}{spectral/ \\non-probabilistic}  &-----  &-----& inconclusive  \\  \hline
B-NR (SBM) & smaller  & probabilistic & very good & moderate & well-fitted    \\  \hline
B-NR (DC-SBM) & smaller & probabilistic  & moderate & very good & over fits, modestly    \\  \hline
B-HKK (SBM) & smaller & probabilistic  & good & moderate & under fits, modestly  \\  \hline
cICL-HKK (SBM) & smaller & probabilistic  & good & moderate & under fits, modestly \\  \hline
Infomap & larger & non-probabilistic  & moderate & moderate & over fits\\  \hline
MDL (SBM) & smaller  & probabilistic  & good & poor & under fits  \\  \hline
MDL (DC-SBM) & smaller  & probabilistic & very good & moderate & well-fitted \\  \hline
S-NB & smaller & spectral  & moderate & moderate & uneven fits \\  \hline
S-cBHm & smaller  & spectral & ----- & ----- & inconclusive   \\   \hline
S-cBHa & smaller  & spectral  & ----- & ----- & inconclusive  \\   \hline
AMOS & larger & non-probabilistic  &----- &-----& inconclusive   \\  \hline
LRT-WB (DC-SBM) & larger  & non-probabilistic &-----&-----& inconclusive \\  \hline \hline
 \label{table:results}
 \end{tabular}
\end{table*}

Figure~\ref{fig:LP_vs_LD} (see Appendix) provides a complementary representation of these results, via a parametric plot (parameterized by $\alpha$), showing accuracy on link prediction as a function of accuracy on link description. The path each method traces through this space illustrates its average tradeoff between prediction and description, as a function of the relative sparsity of the training set. This parametric space may also be partitioned into three regions that align with the evaluation criteria of Box~\hyperref[table:BCD]{3}, 
	so that these regions correspond to good or poor performance on the two tasks (with good performance both being excluded). 

The results in this section reveals substantial evidence that most methods tend to over- or under-fit to networks, to some degree. However, poor performance at either task could also be the result of a poor pairing of a particular score function with the particular communities an algorithm finds. A valuable check on our above results is to test the performance of the identified communities under a common score function. The results of this experiment are available in Appendix~\ref{secA:MA-LPLD}.

\begin{tcolorbox}[breakable,label=table:BCD,float = t!,size=title,
  colback=blue!2!white,colframe=blue!5!black,fonttitle=\bfseries,
  title=Box 3: Behavior of Community Detection methods,pad at break=1mm, break at=-\baselineskip/0pt]
  \begin{tabular}{lp{5.5cm}}
  Well-fitted & link prediction: \hspace{1.5cm} good \\ & link description: \hspace{1.35cm} poor \\ & e.g., MDL and B-NR \vspace{3mm} \\
 Overfitted & link prediction: \hspace{1.5cm} poor \\ & link description: \hspace{1.35cm} good \\ & e.g., Q, Q-MR, and Infomap \vspace{3mm} \\
  Underfitted & link prediction: \hspace{1.5cm} poor \\ & link description: \hspace{1.35cm} poor \\ & e.g., B-HKK and Q-MP \vspace{3mm} \\
  Uneven & overfits on some classes of inputs \\ & underfits on others \\ & e.g., spectral methods \\
  \end{tabular}
\end{tcolorbox} 

\subsection{Discussion of Results}
\label{sec:disc}
One consequence of the No Free Lunch and ``no ground truth'' theorems for community detection~\cite{peel2017ground} is that algorithm evaluations based on comparing against a partition defined by node metadata do not provide generalizable or interpretable results. As a result, relatively little is known about the degree to which different algorithms over- or under-fit on or across different kinds of inputs.

The pair of link-based tasks introduced here defines a tradeoff much like a bias-variance tradeoff, in which an algorithm can either excel at learning the observed network (link description) or at learning to predict missing edges (link prediction), but not both. Link description thus represents a kind of in-sample learning evaluation, and an algorithm with a high benchmark performance curve on this task tends to correctly learn where edges do and do not appear in a given network. In contrast, link prediction presents a kind of out-of-sample learning evaluation, and an algorithm with a high performance curve on this task must do worse at link description in order to identify which observed non-edges are most likely to be missing links. The relative performance on these two tasks provides a qualitative diagnostic for the degree to which an algorithm tends to over- or under-fit when applied to real-world networks. Box~\hyperref[table:BCD]{3} provides simple guidelines for assessing this tendency for any particular method.

Our evaluation and comparison of 11 state-of-the-art algorithms under these two tasks using the CommunityFitNet corpus of 572 structurally diverse networks provides new insights. We summarize our findings in Table~\ref{table:results}, describing the number of communities an algorithm tends to find, the group of algorithms its output partitions tend to be most similar to, and its performance on link prediction and description. We also provide an overall qualitative assessment of the algorithm's practical behavior, following the rubrics in Box~\hyperref[table:BCD]{3}. In particular, we find dramatic differences in accuracy on both tasks across algorithms. Generally, we find that probabilistic methods perform well at link prediction and adequately at link description, indicating that they tend to produce relatively well-fitted communities.

In contrast, we find that methods based on modularity maximization (Q and Q-MR) and Infomap tend to overfit on real-world networks, performing poorly at link prediction but well at link description. In contrast, some methods, such as \mbox{Q-MP} tend to underfit on real-world data, performing poorly, or at best moderately, on both tasks. In previous work on a more limited number of networks, Ref.~\cite{kawamoto2016comparative} concluded that modularity-based spectral community detection tends to underfit, while non-backtracking spectral community detection (S-NB, here) tends to overfit. Our results on 572 networks, however, show that spectral methods such as S-NB are more uneven, tending to underfit in some circumstances and overfit in others. Given the broad popularity of spectral techniques in community detection, this particular result indicates that more careful evaluations of spectral techniques in practical settings are likely warranted, and their general accuracy should not be assumed.

It is worth highlighting that in some settings, an approach like Infomap or Q that is better at link description than at link prediction may be preferred, as overfitting is controlled in a post-processing step outside the algorithm itself. Similarly, some methods, such as Infomap, are more readily adaptable to different kinds of input data and research questions. Articulating and formalizing such qualitative methodological tradeoffs are an important direction for future work.

\begin{figure}[t!]
     \begin{center}
            \includegraphics[width=0.48\textwidth]{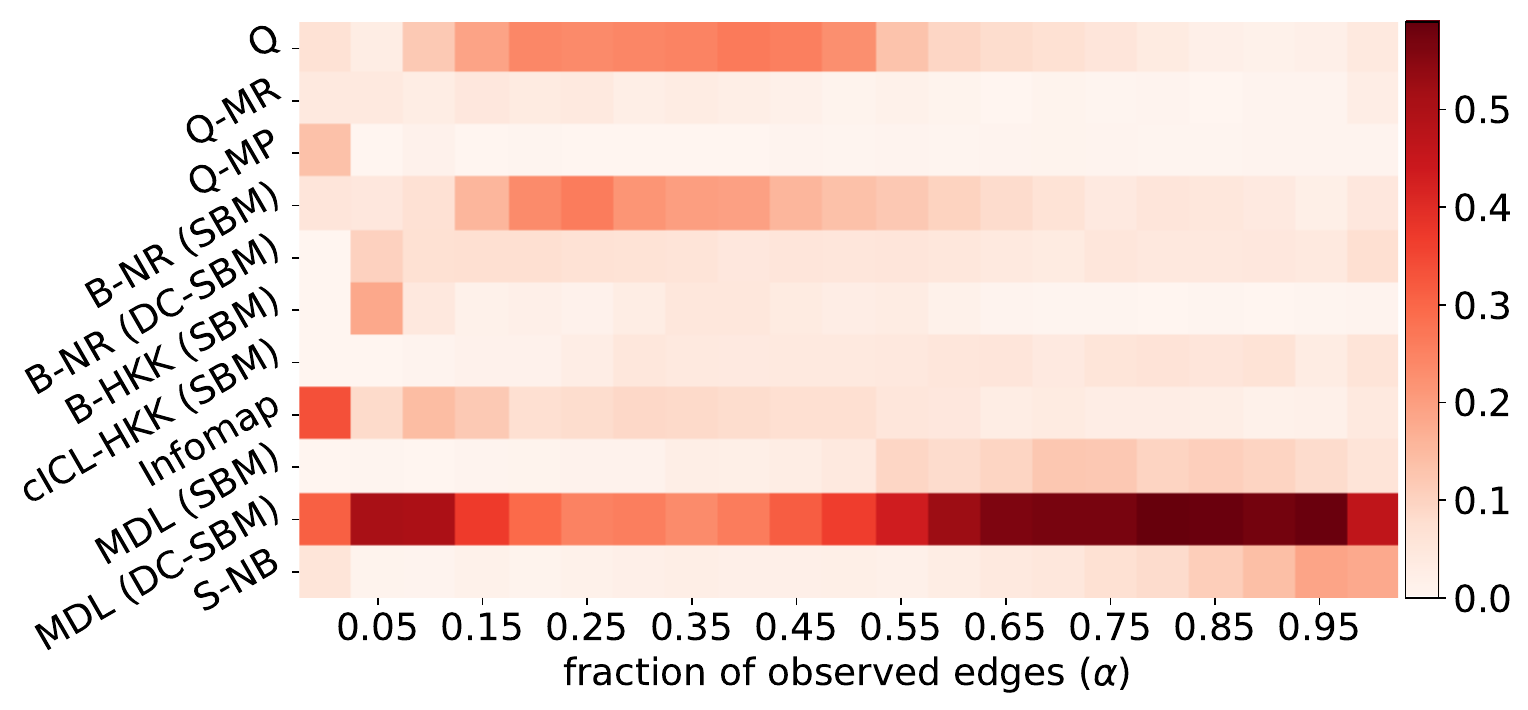}
    \end{center}
    \vspace*{-5mm}
    \caption{A heatmap showing the fraction of networks in the CommunityFitNet corpus on which a particular algorithm produced the best performance on the link prediction task, for different levels of subsampling $\alpha$. 
The two best overall methods (MDL DC-SBM and B-NR SBM) in Fig.~\ref{fig:LP}a are not always the best, and every algorithm is the best for some combination of network and $\alpha$. Here, any  
algorithm with an AUC performance within 0.05 of the maximum observed AUC, for that network and $\alpha$ choice, is also considered to be ``best''.} 

   \label{fig:DBLP}
\end{figure}

The best overall methods are MDL (DC-SBM) and B-NR (SBM). However, their better performance is not universal across the CommunityFitNet corpus, and other algorithms perform better at link prediction on some networks in the corpus than do either of these methods. In fact, we find that \textit{every algorithm} is the best link prediction algorithm for some combination of network and level of subsampling $\alpha$ (Fig.~\ref{fig:DBLP}). In other words, no algorithm is always worse at the link prediction task than every other algorithm, even if there are some algorithms that are on average better than some other algorithms. This variability in the best algorithm for particular networks aligns with the main implication of the No Free Lunch theorem~\cite{peel2017ground}, and illustrates the misleading nature of using evaluations over a small set of real-world networks to support claims that this or that algorithm is generally superior. Our findings demonstrate that such claims are likely incorrect, and that superiority is context specific, precisely as the No Free Lunch theorem would predict.

\begin{figure*}[t!]
     \begin{center}
            \includegraphics[width=0.8\textwidth]{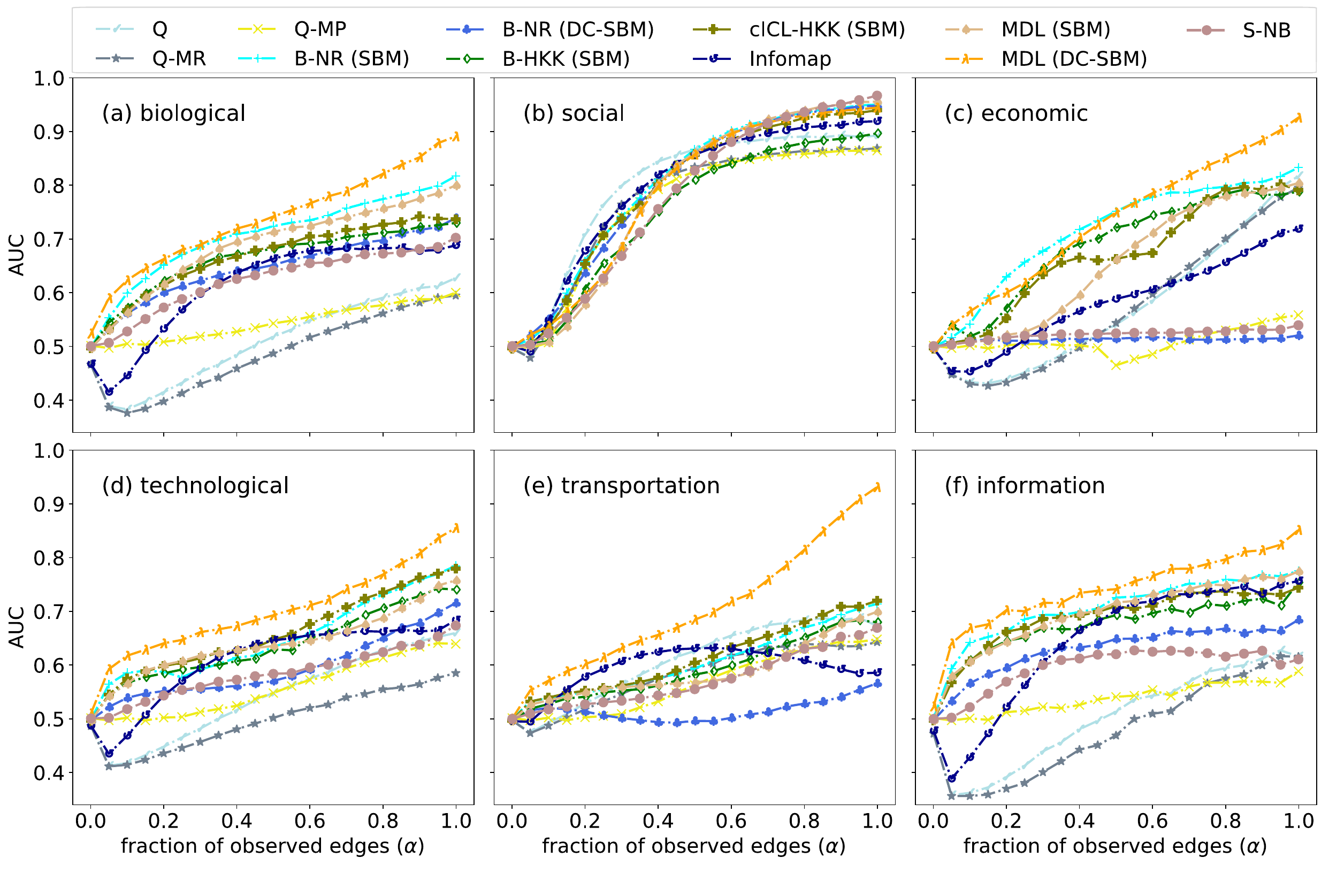}
    \end{center}
    \vspace*{-5mm}
    \caption{Separate benchmark performance curves using model-specific score functions for the link prediction (test) task for networks drawn from (a) biological (34\%), (b) social (22\%), (c) economic (21\%), (d) technological (12\%), (e) transportation (7\%), and (f) information (4\%) domains of origin in the CommunityFitNet corpus. As in Fig.~\ref{fig:LP}a,  each curve shows the mean AUC for a different community detection method, for a given fraction $\alpha$ of observed edges in a network.}
   \label{fig:LP_domain}
\end{figure*}

Reinforcing this point, we also find that algorithm performance on the link prediction task varies considerably depending on the type of network, i.e., whether the network is drawn from a social, biological, economic, technological, transportation, or information domain (Fig.~\ref{fig:LP_domain}). Remarkably, nearly all methods perform similarly well on social networks, which may reflect the tendency for social networks to exhibit simple assortative mixing patterns---communities with many more edges within them than between them---that are well-captured by every algorithm. In contrast, networks from other domains present more variable results, sometimes markedly so. Technological networks produce more modest performance across all algorithms, but with more cross-algorithm variability than observed in social networks, and both performance and variability are greater still for information and biological networks. The greatest variability is seen for economic, information, and biological networks, which suggests the presence of structural patterns not well-captured by the poor-performing algorithms. 

A thorough exploration of the reasons that some algorithms perform more poorly in some domains than others would be a valuable direction for future work. 
One candidate explanation comes from the prevalence or presence of 
disassortative communities---communities defined more by the absence of their interconnections than their presence---in networks. Many community detection algorithms are designed primarily to find assortative communities, and hence the presence of disassortative patterns may cause overfitting in their output. To test this simple hypothesis, we separated our results into two classes, depending on whether they came from a bipartite network, and hence one that is naturally disassortative, or a non-bipartite network. These results, given in Appendix~\ref{secA:BPvsNBP}, show that algorithms exhibit the same qualitative patterns of overfitting and underfitting for both classes. We look forward to a deeper investigation of the origins of this behavior in future work.

\section{Conclusion}
\label{sec:conc}
Community detection algorithms aim to solve the common problem of finding a lower-dimensional description of a complex network by identifying statistical regularities or patterns in connections among nodes. However, no algorithm can solve this problem optimally across all inputs~\cite{peel2017ground}, which produces a natural need to characterize the actual performance of different algorithms across different kinds of realistic inputs. Here, we have provided this characterization, for 16 state-of-the-art community detection algorithms applied to a large and structurally diverse corpus of real-world networks. 
The results shed considerable light on the broad diversity of behavior that these algorithms exhibit when applied to a consistent and realistic benchmark.

For instance, nearly all algorithms appear to find a number of communities that scales like $O(\sqrt{M})$. At the same time, however, algorithms can differ by more than an order of magnitude in precisely how many communities they find within the same network (Fig.~\ref{fig:AD_NE}). And, non-probabilistic approaches typically find more communities than probabilistic approaches. Comparing the precise composition of the identified communities across algorithms indicates that algorithms with similar underlying assumptions tend to produce similar kinds of communities---so much so that we can cluster algorithms based on their outputs (Fig.~\ref{fig:dend-NMI}), with spectral techniques finding communities that are more similar to those found by other spectral techniques than to communities found by any other methods, and similarly for probabilistic methods and for non-probabilistic methods. This behavior would seem to indicate that different sets of reasonable assumptions about how to specify and find communities tend to drive real differences in the composition of the output. That is, different assumptions reflect different underlying tradeoffs, precisely as predicted by the No Free Lunch theorem.

Different algorithms also present wide variation in their tendency to over- or under-fit on real networks (Fig.~\ref{fig:LP}), and the link prediction/description tasks we introduced provide a principled means by which to characterize this algorithmic tendency. Here also we find broad diversity across algorithms, with some algorithms, like MDL (DC-SBM) and B-NR (SBM) performing the best on average on link prediction and well enough on link description. However, we also find that these algorithms are not always the best at these tasks, and other algorithms can be better on specific networks (Fig.~\ref{fig:DBLP}). This latter point is cautionary, as it suggests that comparative studies of community detection algorithms, which often rely on a relatively small number of networks by which to assess algorithm performance, are unlikely to provide generalizable results. The results of many previously published studies may need to be reevaluated in this light, and future studies may find the link prediction and link description tradeoff to be a useful measure of algorithm performance.

Beyond these insights into the algorithms themselves, the CommunityFitNet corpus of networks has several potential uses, e.g., it can be used as a standardized reference set for comparing community detection methods. To facilitate this use case, both the corpus dataset and the derived partitions for each member network by each of the algorithms evaluated here is available online for reuse. To compare a new algorithm with those in our evaluation set, a researcher can simply run the new algorithm on the corpus, and then identify which reference algorithm has the most similar behavior, e.g., in the average number of communities found (Fig.~\ref{fig:AD_NE}) or the composition of the communities obtained (Fig.~\ref{fig:dend-NMI}). Similarly, a researcher could quickly identify specific networks for which their algorithm provides superior performance, as well as compare that performance on average across a structurally diverse set of real-world networks. We expect that the availability of the CommunityFitNet corpus and the corresponding results of running a large number of state-of-the-art algorithms on it will facilitate many new and interesting advances in developing and understanding community detection algorithms.

Our results also open up several new directions of research in community detection. For instance, it would be valuable to investigate the possibility that a method, when applied to a single network, might over-partition some parts but under-partition other parts---an idea that could be studied using appropriate cross-validation on different parts of networks. Similarly, a theoretical understanding of what algorithmic features tend to lead to over- or under- or uneven-fitting outcomes for community detection would shed new light on how to control the underlying tradeoffs that lead to more general or more specific behavior. These tradeoffs must exist~\cite{peel2017ground}, and we find broad empirical evidence for them across our results here, but there is as yet no theoretical framework for 
understanding what they are or how to design around them for specific network analysis or modeling
tasks.

\appendices

\begin{figure*}[t!]
     \begin{center}
            \includegraphics[width=1\textwidth]{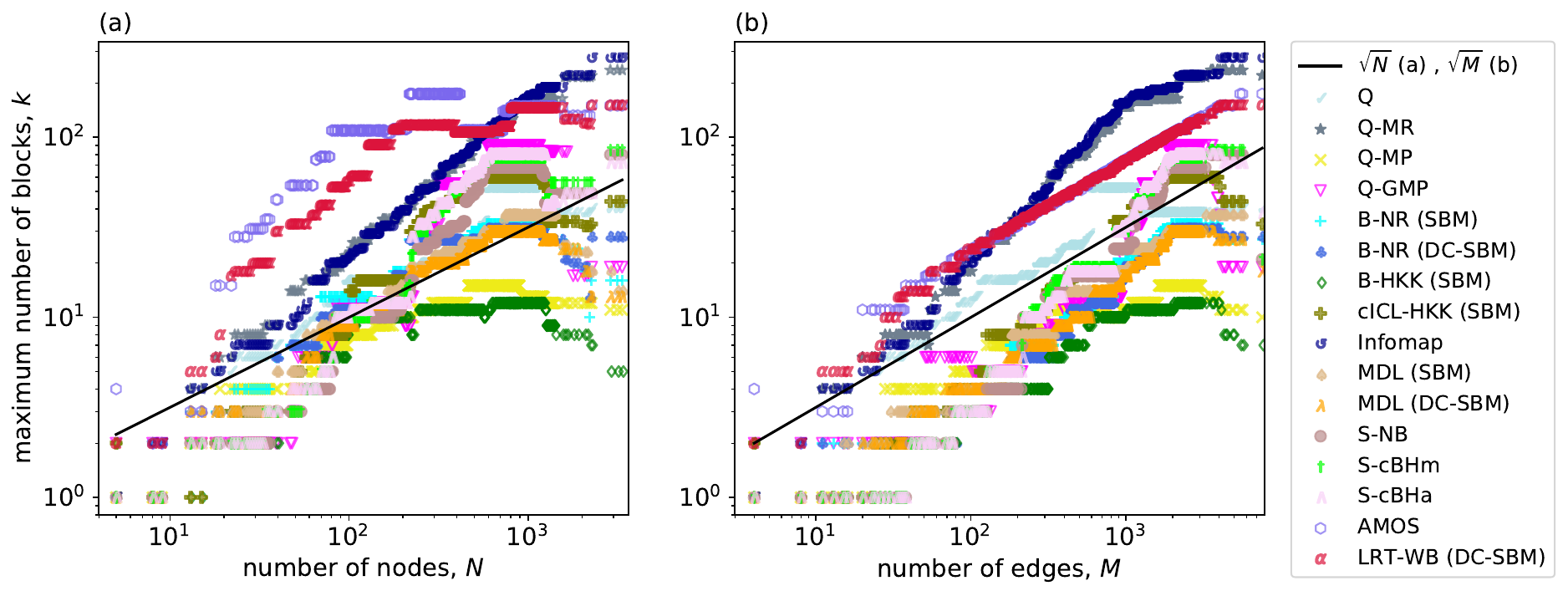}
    \end{center}
    \vspace*{-5mm}
    \caption{The maximum number of inferred communities, for 16 state-of-the-art methods (see Table 1) applied to 572 real-world networks from diverse domains, versus the (a) number of nodes $N$, with a theoretical prediction of $\sqrt{N}$, or (b) number of edges $M$, with a theoretical prediction of $\sqrt{M}$.}%
   \label{fig:MD_NE}
\end{figure*}
\section{Performance On Bipartite versus Non-Bipartite Networks}
\label{secA:BPvsNBP}
\begin{figure*}[t!]
     \begin{center}
            \includegraphics[width=1\textwidth]{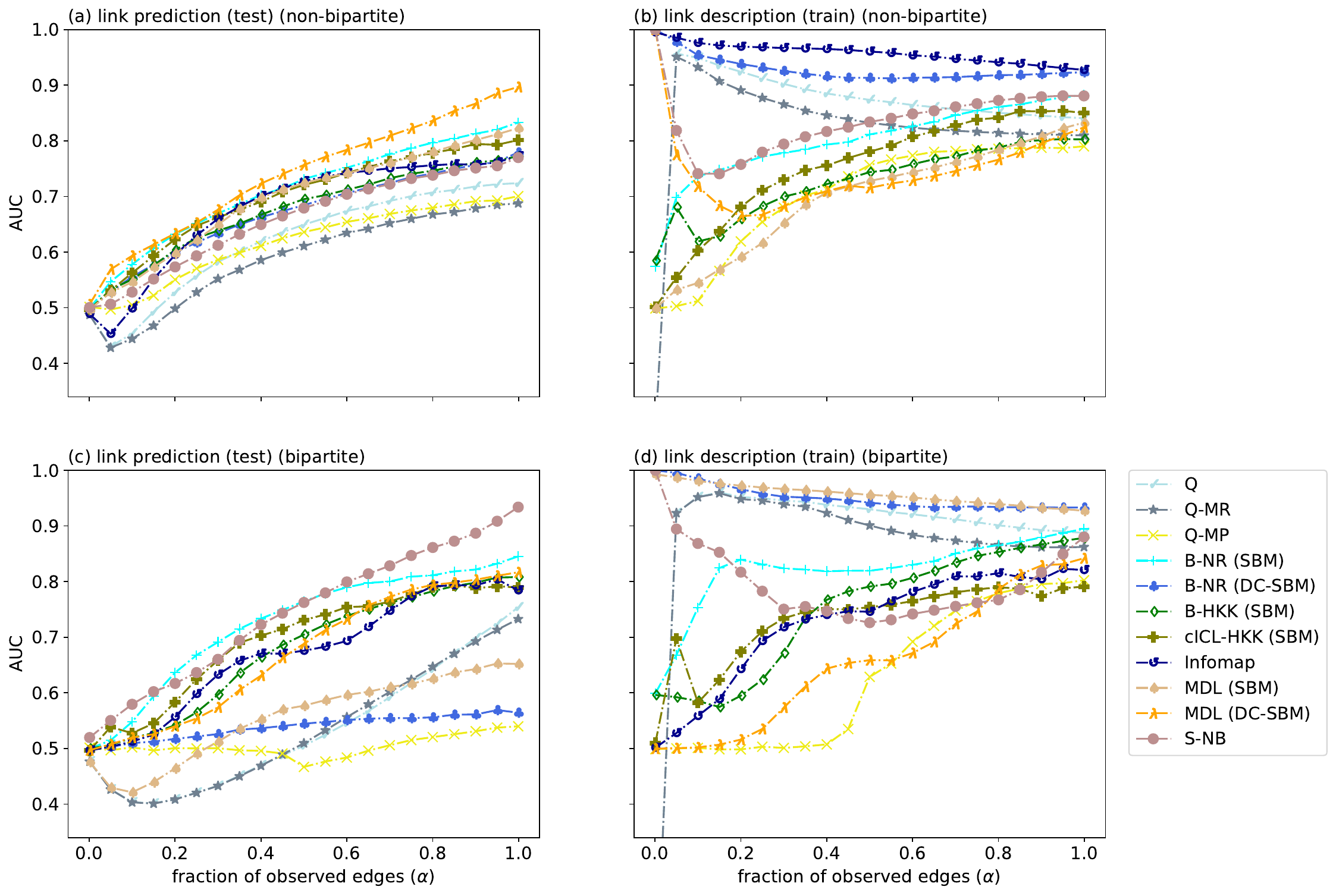}
    \end{center}
    \vspace*{-5mm}
    \caption{Separate benchmark performance curves using model-specific score functions for link prediction and link description tasks for networks drawn from ({\it top}) non-bipartite (73\%), ({\it bottom}) bipartite (27\%) networks of origin in the CommunityFitNet corpus. As in Fig.~\ref{fig:LP}a,  each curve shows the mean AUC for a different community detection method, for a given fraction $\alpha$ of observed edges in a network.}
   \label{fig:LPLD_BPvsNBP}
\end{figure*}

In this section, the networks are categorized into two groups of bipartite networks and non-bipartite networks to understand how this characteristic affects the performance among our set of community detection algorithms. The summary statistics of the CommunityFitNet corpus for these two groups of networks are provided for different domains in Table~\ref{table:summary_stat}. The main goal of this section is to explore whether the bipartite networks cause overfitting in algorithms like Infomap and modularity variants like Q and Q-MR. Fig.~\ref{fig:LPLD_BPvsNBP} presents the link prediction and link description for each category. Similar patterns of overfitting and underfitting can still be seen in non-bipartite networks. An interesting pattern is revealed by comparing \mbox{MDL (DC-SBM)} and \mbox{B-NR (SBM)}: \mbox{B-NR (SBM)} has partially better performance on bipartite networks compared to \mbox{MDL (DC-SBM)}. In contrast, the performance of MDL (DC-SBM) is slightly better in non-bipartite networks. 

\begin{table*}[t!]\addtolength{\tabcolsep}{-5pt}
\caption{The summary statistics of CommunityFitNet corpus in each domain for bipartite versus non-bipartite networks.\newline The numbers show (number of non-bipartite)/(number of bipartite) networks.}
\vspace{-3mm}
\centering
 \begin{tabular}{p{2.5cm}p{2.5cm}p{2.5cm}p{2.5cm}p{2.5cm}p{2.5cm}p{2.5cm}} 
 \hline
 Social & Economic & Biological & Technological & Information & Transportation & Total  \\ [0.5ex] 
 \hline\hline
 123/1&11/111&147/45&71/0&22/0&41/0&415/157 \\  \hline \hline
 \label{table:summary_stat}
 \end{tabular}
\end{table*}

\begin{figure*}[t!]
     \begin{center}
            \includegraphics[width=0.8\textwidth]{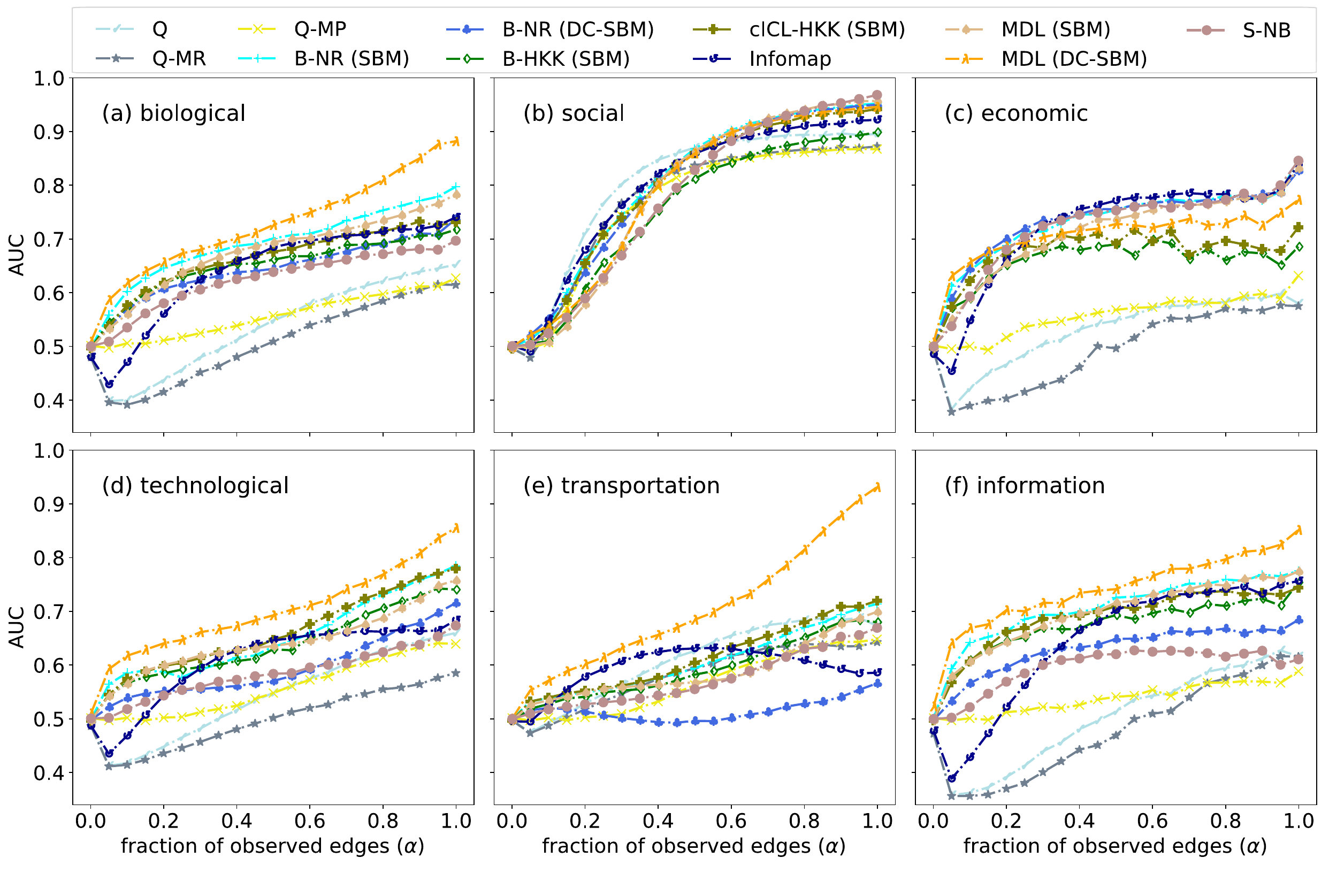}
    \end{center}
    \vspace*{-5mm}
    \caption{Separate benchmark performance curves using model-specific score functions for the link prediction (test) task for non-bipartite networks drawn from (a) biological (35\%), (b) social (30\%), (c) economic (3\%), (d) technological (17\%), (e) transportation (10\%), and (f) information (5\%) domains of origin in the CommunityFitNet corpus. As in Fig.~\ref{fig:LP}a,  each curve shows the mean AUC for a different community detection method, for a given fraction $\alpha$ of observed edges in a network.}
   \label{fig:LP_nbp_domain}
\end{figure*}

Although the link prediction of non-bipartite networks seems to be less variable, this is because almost all social networks are non-bipartite causing the average to follow the trend in this group. However, as mentioned earlier, the same patterns of overfitting and underfitting can be seen in the corpus of non-bipartite networks. This behavior is revealed when we look at the domain separated figure of link prediction for non-bipartite networks, Fig.~\ref{fig:LP_nbp_domain}. 

As it can be seen from Fig.~\ref{fig:LP_nbp_domain}, the overfitting still exists in the methods of modularity and Infomap for technological, biological, and transportation networks. Also Q and Q-MR overfits for economic and information networks. The MDL (DC-SBM) is almost the best algorithms in all domains for non-bipartite networks, especially in transportation networks. B-NR (SBM) is also one of the best compared to other algorithms. Generally Infomap has better predictive performance compared to Q and Q-MR for biological, economic, technological, and information networks. However, Q and Q-MR has better link prediction in transportation networks when the edge density in the observed network is high enough ($\alpha > 0.5$). The spectral method and B-NR (DC-SBM) are among the best methods for non-bipartite economic networks, while they behave almost as poorly as random guessing on average for mixed of bipartite and non-bipartite economic networks.

\section{Performance Under a Common Score Function}
\label{secA:MA-LPLD}

Comparing link prediction and link description benchmark performance curves of 11 state-of-the-art community detection methods reveals substantial evidence that most methods tend to over- or under-fit to networks, to some degree. However, poor performance at either task could also be the result of a poor pairing of a particular score function with the particular communities an algorithm finds.

A valuable check on our above results is to test the performance of the identified communities under a common score function. This experiment thus serves to remove differences in the way the various scoring functions utilize the same partition structure. Specifically, we repeat both link prediction and description evaluation tasks, using the community partitions identified by each of the 11 algorithms for each network in the corpus, and then applying the SBM score function from Section~\ref{sec:MS-LPLD} to construct the benchmark performance curves. Although any score function could be used as such a reference point, the SBM score function has the attractive property that it yielded high general performance for link prediction. This comparison also can be helpful as a sanity check to see if the proposed score functions are good choices for link prediction to test the generalizability performance of the community detection methods. For example, if the algorithm does poorly under its own score function, but well under the SBM score function, then it implies that its own score function is the cause of its poor performance. However, for most of these choices, the chosen score function is the reasonable choice corresponding to the community detection algorithm.

\subsection*{Results} \label{sec:MA-R}

\begin{figure*}[t!]
     \begin{center}
            \includegraphics[width=1\textwidth]{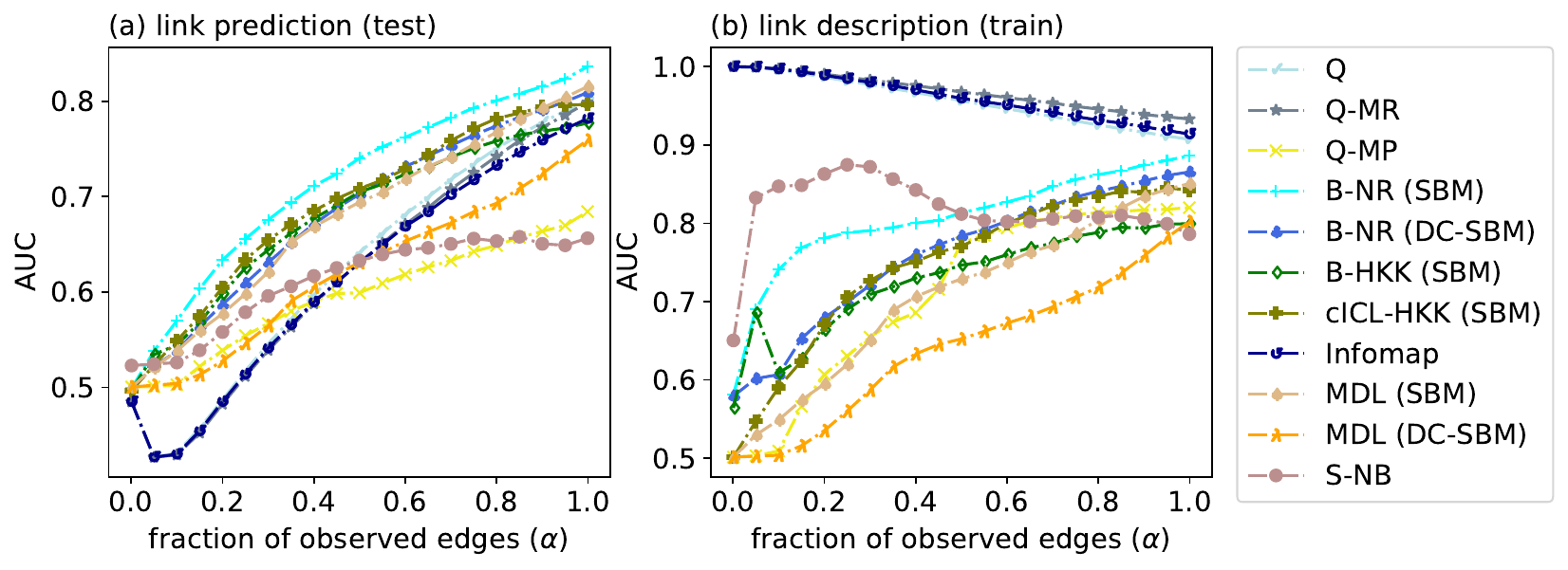}
    \end{center}
    \vspace*{-5mm}
    \caption{Benchmark performance curves using a SBM-based score function for (a) link prediction and (b) link description tasks. Each curve shows the mean AUC for a different community detection method across 572 real-world networks for a given fraction $\alpha$ of observed edges in a network.}
   \label{fig:LF}
\end{figure*}

The relative ordering of the benchmark performance curves under the common score function for the link prediction and description evaluations (Fig.~\ref{fig:LF}) differs in interesting ways from that of the model-specific evaluation (Fig.~\ref{fig:LP}). We note that the performance curves for the SBM-based methods are unchanged as their score function is the same in both settings.

In link prediction, the previous  performance gap between the \mbox{B-NR} (DC-SBM) and \mbox{B-NR} (SBM) methods is much smaller, which shows that the poor performance of \mbox{B-NR} (DC-SBM) in Section~\ref{sec:MS-LPLD} is due to its score function. The \mbox{B-NR} is now the best overall method by a sizable margin and the MDL (DC-SBM) method that produced the best model-specific results for link prediction, performs substantially worse under the SBM-based score function on both tasks. Of course, SBM-based methods should produce communities that exhibit better performance under an SBM-based score function than would other methods. But the DC-SBM in particular was originally designed to find more reasonable communities than the SBM, by preventing the model from fitting mainly to the network's degree structure~\cite{karrer2011stochastic}. The worse performance by the DC-SBM communities on link prediction in this setting indicates these methods' allowance of a lower
entropy in the inferred block structure acts to over-regularize the communities from the perspective of the SBM. Furthermore, unlike the SBM score function, the MDL (DC-SBM) score function (used in Fig.~\ref{fig:LP}) depends on the model complexity, the inclusion of which evidently serves to improve link predictions at all values of $\alpha$. However, link prediction using the inferred communities alone appears to be a slightly unfair evaluation of the DC-SBM (also suggested by Ref.~\cite{aicher2015learning}).

Turning to other methods, recall that Infomap and Q-MR found similar numbers of communities and had similar accuracies in the model-specific link prediction task (Fig.~\ref{fig:LP}). Under the common SBM-based score function, we find that Infomap, Q-MR, and Q exhibit nearly identical performance on both link prediction and description tasks. In light of our previous discussion of the tendency of modularity-based methods to overfit, this similarity, which must derive from these methods all identifying similar community structures in networks, provides additional evidence that all three methods tend to overfit real data.

Finally, the S-NB method shows unusual behavior: in link prediction, its performance is similar under both score functions; and, in link description, its performance under its own model-specific score function is replaced with a non-monotonic performance curve, which is better at lower values of $\alpha$ than at higher values.
The behavior at smaller values of $\alpha$, when the sampled networks are relatively more sparse, is consistent with a tendency for S-NB to under-fit in this regime, in agreement with past results that suggest that spectral methods tend to under-fit when communities are unbalanced~\cite{le2015estimating}. However, the change at larger values of $\alpha$ indicates that, as more edges are sampled, this spectral technique qualitatively changes in its behavior.
Recall that the maximum number of clusters inferred by spectral methods for large networks exceeds the theoretical bound (Fig.~\ref{fig:MD_NE}), which indicates a tendency to overfit. Hence, the relatively worse performance at larger values of $\alpha$ on both tasks suggests that spectral techniques behave differently across different settings, overfitting in large sparse networks, underfitting when communities are unbalanced, and ``well-fitting'' when communities are balanced.
An algorithm that exhibits this kind of context-dependent behavior is deemed to exhibit an ``uneven'' fit. Excluding the non-monotonic performance curve of link description for \mbox{S-NB}, the general comparison between SBM-based and model specific performance shows that the proposed score function for this algorithm is a reasonable choice and is not the reason for the observed poor performance of this algorithm. \\

\begin{figure*}[t!]
     \begin{center}
            \includegraphics[width=0.7\textwidth]{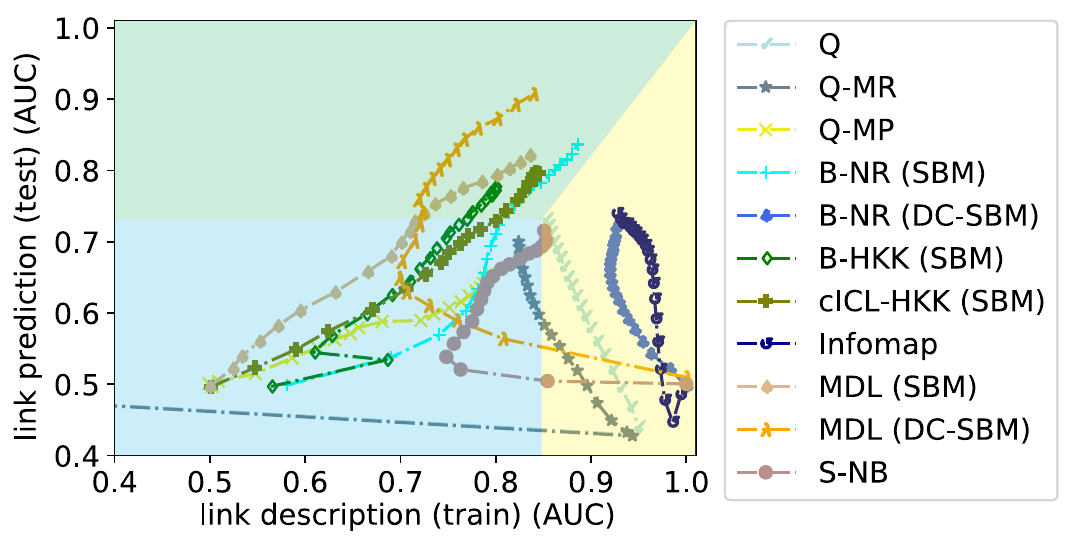}
    \end{center}
    \vspace*{-5mm}
    \caption{A parametric plot showing link prediction versus link description performance, with $\alpha$ parameterizing the trajectory of each line.}
   \label{fig:LP_vs_LD}
\end{figure*}

\section{other representations of link prediction and link description}
As described in the main text, the larger/smaller number of clusters a community detection algorithm finds is consistent with the formal definitions of overfitting/underfitting in non-relational data. The link prediction and link description definitions are conceptually similar to prediction on the test set and training set, respectively. This relationship recalls the bias-variance tradeoff in traditional machine learning, where increasing the model complexity decreases the training error and increases the test error. The link prediction and link description performance curves are very similar to these plots where the model complexity is replaced with the number and composition of communities found. 

Another useful representation for these tasks is by combining them in a parametric plot with parameter $\alpha$. In Fig.~\ref{fig:LP_vs_LD}, we divided the performance space into three different regions that roughly correspond to good-poor, poor-good, and poor-poor (link prediction-link description) performance. As shown in Box~\hyperref[table:BCD]{3}, these regions correspond to well-fitted, overfitted, and underfitted behaviors of community detection algorithms, respectively. 
\section{Scoring function}
In this section, for reproducibility of our results, we will explain in detail the scoring functions we used. Also we will explain additional details of the link prediction and link description procedures for these algorithms. For running time considerations, as mentioned in the main text, we approximate the $AUC$ via the Monte Carlo sampling.
\subsection{\mbox{B-NR (SBM)}, \mbox{B-NR (DC-SBM)}, \mbox{B-HKK}, \mbox{cICL-HKK}, and \mbox{S-NB}}
This group of methods have the characteristic that the value assigned to each pair of nodes doesn't depend on the existence of the link. The natural score function for each pair of nodes defined for the probabilistic methods (\mbox{B-NR (SBM)}, \mbox{B-NR (DC-SBM)}, \mbox{B-HKK}, and \mbox{cICL-HKK}) is the probability of the existence of the corresponding query edge~\cite{guimera2009missing}. For spectral clustering \mbox{S-NB}, as explained in Section~\ref{sec:MS-LPLD}, a new score function based on eigenvalue decomposition is constructed. The proposed spectral scoring rule $s_{ij}$ is the corresponding entry value in the low-rank approximation with the rank coming from the non-backtracking spectral method.
\subsubsection{Link Prediction}
For link prediction for these methods, we compare the pairwise scores for missing links and non-links to compute the $AUC$ using Monte Carlo sampling. We remove $(1-\alpha) \%$ of the links uniformly, randomly choose 10000 pairs of missing links and non-links, and compare the scores on pairs of missing links and non-links to compute the $AUC$.
\subsubsection{Link Description}
For link description, we compare the pairwise scores for links and non-links to compute the $AUC$ using Monte Carlo sampling. We remove $(1-\alpha) \%$ of the links uniformly, randomly choose 10000 pairs of observed links and non-observed links, and compare the scores on pairs of observed links and non-observed links to compute the $AUC$.
\subsection{Q, Q-MR, Q-MP, Infomap, MDL (SBM), and MDL (DC-SBM)}
Here, we summarize the score functions for the non-probabilistic score function methods~\footnote{Some of these methods, like \mbox{MDL (SBM)} and \mbox{MDL (DC-SBM)} are closely related to probabilistic methods but their score functions are non-probabilistic.}. The score function of each potential edge $i,j$ for each of these algorithms is defined as the amount of contribution that query edge makes in the corresponding objective function. For example in modularity \mbox{Q}, the objective function is computed as $Q=\sum_{r=1}^K \left[\dfrac{l_r}{M} - \left(\dfrac{d_r}{2M}\right)^2 \right]$~\cite{newman2006modularity}, where $l_r$ is the number of edges inside group $r$, $d_r$ is the aggregated degree of nodes of type $r$, and $M$ is total number of edges in the network. The score function $s_{ij}$ for a query edge $i,j$ is the increase in modularity $\Delta Q$, after adding that edge into the network. For running time considerations, we assume the partitions remain unchanged after adding the edge. 
\subsubsection{Link Prediction}
For link prediction, we remove $(1-\alpha) \%$ of the links uniformly, randomly choose 10000 pairs of missing links and non-links, then once add a link in the location of the missing link, and once add a link in the location of the non-link and see whose contribution is larger to compute the $AUC$ (see Fig.~\ref{fig:LPLD_com_scheme}(a)).  
\subsubsection{Link Description}
\begin{figure*}[t!]
     \begin{center}
            \includegraphics[width=0.5\textwidth]{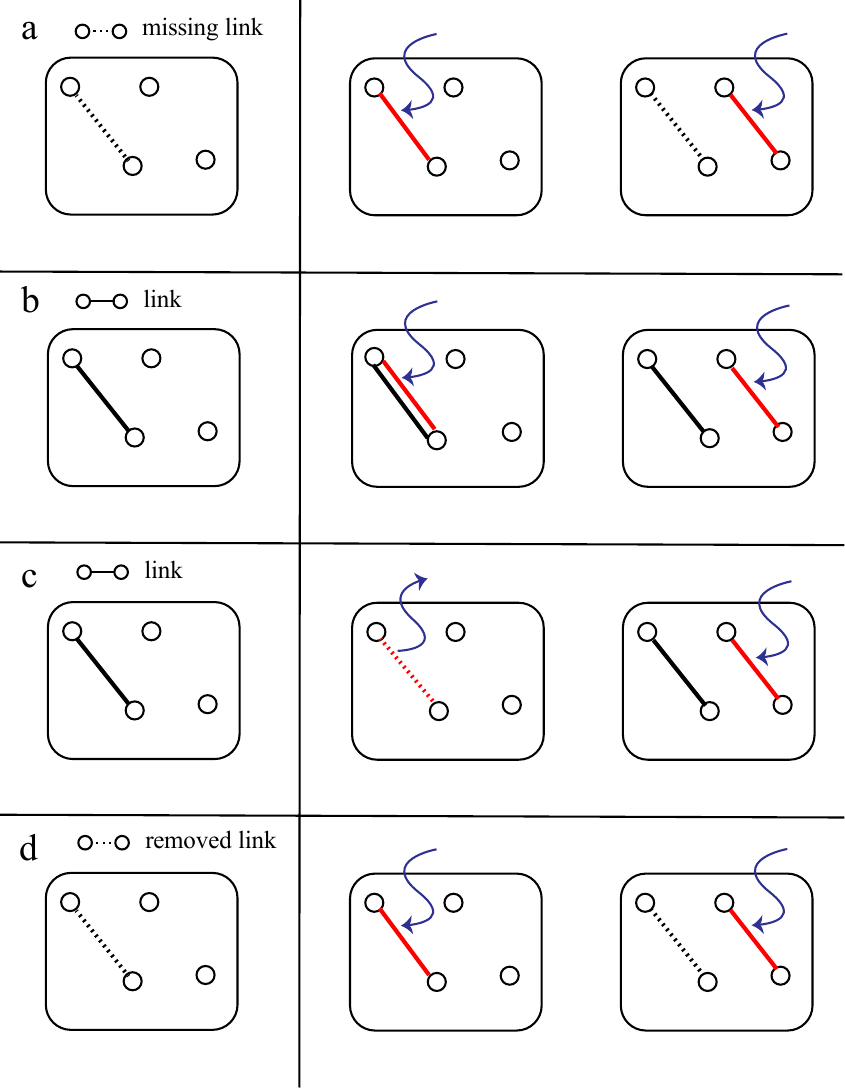}
    \end{center}
    \vspace*{-5mm}
    \caption{Computation of (a) link prediction versus (b,c,d) link description in non-probabilistic score function methods of \mbox{Q}, \mbox{Q-MR}, \mbox{Q-MP}, \mbox{Infomap}, \mbox{MDL (SBM)}, and \mbox{MDL (DC-SBM)}.  (a) Consider the current network as the reference, once add a link in the location of the missing link, and once add a link in the location of the non-link and see whose contribution is larger to compute the $AUC$, (b) consider the current network as the reference, once add a link in the location of the link and once add a link to the location of the non-link and see whose contribution is larger in the objective function to compute the $AUC$, (c) consider the current network as the reference, once remove a link from the location of the link and once add a link in the location of the non-link to see whose contribution is larger to compute the $AUC$, and (d) remove the link and consider it as the reference, once add a link in the location of the removed link, and once add a link in the location of the non-link to see whose contribution is larger to compute the $AUC$.}
   \label{fig:LPLD_com_scheme}
\end{figure*}

In link description, we remove $(1-\alpha) \%$ of the links uniformly, and randomly choose 10000 pairs of observed links and non-observed links. Here, we have three different options to compare the contribution of these two groups (see Fig.~\ref{fig:LPLD_com_scheme}). 

In Fig.~\ref{fig:LPLD_com_scheme} (b), we consider the current network as the reference, then once add a link in the location of the link and add a link to the location of the non-link and see whose contribution is larger in the objective function to compute the $AUC$. The reason of performing the link description this way is we want the learned model to automatically find the position of the links without prior knowledge. 

In two other cases, Fig.~\ref{fig:LPLD_com_scheme} (c) and (d), we assumed the learned model knows the position of the links. In Fig.~\ref{fig:LPLD_com_scheme} (c), we consider the current network as the reference, once remove a link from the location of the link and once add a link in the location of the non-link to see whose contribution is larger to compute the $AUC$. And finally in Fig.~\ref{fig:LPLD_com_scheme} (d), we remove the link and consider it as the reference, once add a link in the location of the removed link, and once add a link in the location of the non-link to see whose contribution is larger to compute the $AUC$. 

\begin{figure*}[t!]
     \begin{center}
            \includegraphics[width=1\textwidth]{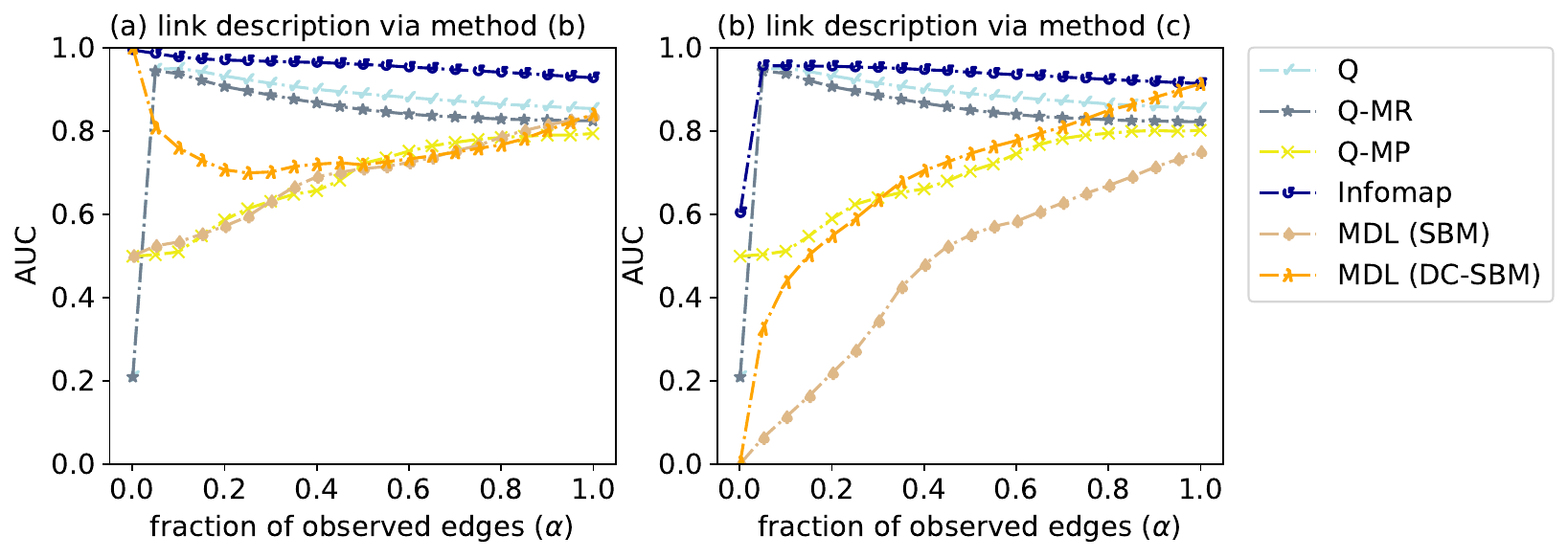}
    \end{center}
    \vspace*{-5mm}
    \caption{Comparison of link description (train) benchmark performance curves for non-probabilistic score function methods of \mbox{Q}, \mbox{Q-MR}, \mbox{Q-MP}, \mbox{Infomap}, \mbox{MDL (SBM)}, and \mbox{MDL (DC-SBM)} using method (b) versus method (c) in comparing contribution of observed links versus non-observed links in link description.}
   \label{fig:LD_b_vs_c}
\end{figure*}

The link description results for these methods, presented in the manuscript, are through using the method (b). Fig.~\ref{fig:LD_b_vs_c} compares the results for method (b) and method (c). Although the results do change slightly, our main conclusions are not affected when using method (c) in computing the contribution of the observed links versus non-observed links. Also is worth highlighting that comparison using method (d) is very computationally expensive; since for every pair after removing the true link for reference, we have to run the algorithm again which adds a large time complexity compared to options (b) and (c).

\section{Model Selection Approaches}
\label{secA:MSA}

The general problem of choosing the number of communities $k$ is a kind of model selection problem, specifically, a kind of complexity control, as selecting more communities generally means more flexibility for the fitted model.  Although it may be appealing to attempt to divide approaches based on whether $k$ is chosen explicitly as a parameter (as in many probabilistic approaches, like the SBM and its variants) or implicitly as part of the community detection itself (as in modularity maximization), such a dichotomy is not particularly clean. In this section, we survey the various different approaches to model selection in community detection. 

Community detection methods can be divided in two broad categories: probabilistic and non-probabilistic methods. These two general groups cover roughly six classes of methods:
\begin{itemize}
\item Bayesian marginalization and regularized likelihood\footnote{The frequentist approaches belong to the regularized likelihood approaches.} approaches~\cite{hofman2008bayesian,yan2016bayesian,daudin2008mixture}, 
\item information theoretic approaches~\cite{rosvall2008maps},  
\item modularity based methods~\cite{newman2004finding}, 
\item spectral and other embedding techniques~\cite{Krzakala2013,saade2014spectral,le2015estimating,grover2016node2vec,perozzi2014deepwalk},
\item cross-validation methods~\cite{chen2014network,kawamoto2016cross}, and
\item statistical hypothesis tests~\cite{wang2015likelihood}.
\end{itemize}
We note, however, that the boundaries among these classes are not rigid and one method can belong to more than one group. For example, MDL is both an information theoretic approach as well as a Bayesian approach, and modularity can be viewed as a special case of the DC-SBM.

Many probabilistic approaches choose a parametric model like the popular SBM or one of its variants, and then design specific rules for model selection (choosing $k$) around this basic model. One principled way to avoid overfitting is to use the minimum description length (MDL)~\cite{rissanen1978modeling} method, which tries to compress the data via capturing its regularities. Ref.~\cite{peixoto2013parsimonious} employs MDL on networks and aims to avoid overfitting via trading off the goodness of fit on the observed network with the description length of the model. This approach can also be generalized to hierarchical clustering and overlapping communities~\cite{peixoto2014hierarchical,peixoto2015model}.

The probabilistic group includes the Regularized-Likelihood approaches~\cite{daudin2008mixture,latouche2012variational,come2015model} and Bayesian model selection methods~\cite{hofman2008bayesian,newman2016estimating,yan2016bayesian,hayashi2016tractable}. 
Regularized-Likelihood approaches are similar to Bayesian Information Criterion (BIC) in model selection~\cite{schwarz1978estimating}. 
Ref.~\cite{biernacki2000assessing} proposes to select the number of clusters in mixture models, using some criterion called the integrated complete likelihood (ICL) instead of BIC. Basically, BIC does not take into account the entropy of the fuzzy classification and ICL is intended to find more reliable clusters by adding this entropy into the penalty terms. However, computing ICL in the setting of a network mixture model like the SBM is not tractable.
To address this issue for the SBM, Ref.~\cite{daudin2008mixture} proposed using ICL and approximating it by resorting to the frequentist variational EM. 
Because of asymptotic approximations in ICL, these results are not reliable for smaller networks. In another study~\cite{latouche2012variational}, the authors employ variational Bayes EM and propose to use the ILvb criterion for complexity control. In both the ICL and ILvb~\cite{daudin2008mixture,latouche2012variational}
approaches, some approximations are used. Ref.~\cite{mcdaid2013improved} bypasses these approximations by considering the conjugate priors and tries to improve the results by finding an analytical solution. Also in Ref.~\cite{come2015model}, the authors find the exact ICL by using an 
analytical expression and propose a greedy algorithm to find the number of clusters and partition the network simultaneously.

We categorize regularized-likelihood and Bayesian approaches together, because the prior beliefs in Bayesian approaches play a similar role to penalty terms in penalized likelihood functions. Bayesian marginalization and related approaches aim to control for overfitting by averaging over different parameterizations of the model. The various approaches in the Bayesian group use different approximations in order to make this averaging computationally feasible in a network setting. 
A common practice for networks, e.g., starting with the SBM, is to either use a Laplace approximation or use conjugate priors~\cite{newman2016estimating,yan2016bayesian,hayashi2016tractable}, both of which yield a penalty term that can be compared with penalty terms in regularized methods. 
Different choices in the particular priors~\cite{newman2016estimating,yan2016bayesian} or in the 
order of Laplace approximations~\cite{newman2016estimating,hayashi2016tractable} yield different resulting model selection specifications. Similarly, Ref.~\cite{newman2016estimating} chooses a maximum entropy prior (B-NR), while Ref.~\cite{yan2016bayesian} chooses a uniform prior.

An approximation technique known as factorized information criterion (FIC) is explored in the context of networks in Ref.~\cite{hayashi2016tractable}, along with its corresponding inference method known as factorized asymptotic Bayesian (FAB). Ref.~\cite{hayashi2016tractable} adapts this criterion to the 
SBM and name it $\mt{F}^2\mt{IC}$, which is more precise than FIC and is specifically designed for SBM. A tractable algorithm named $\mt{F}^2\mt{AB}$ (B-HKK) is proposed to carry out Bayesian inference. A key property of the FIC is that it can account for dependencies among latent variables and parameters, and is asymptotically consistent with the marginal log-likelihood. 
Ref.~\cite{hayashi2016tractable} also proposes a modification to the ICL criterion~\cite{daudin2008mixture} that corresponds to the simplified version of FIC~\cite{hayashi2015rebuilding}, and which is referred to as corrected ICL (cICL-HKK) here.

In contrast to the description length approaches taken with probabilistic models like the SBM, Ref.~\cite{rosvall2008maps} proposes a different information theoretic approach known as Infomap, which uses compression techniques on the paths of a random walker to identify community structure regularities in a network. This approach can be generalized to hierarchical community structure~\cite{rosvall2011multilevel} and to overlapping modular organization~\cite{esquivel2011compression}.

In modularity based methods~\cite{newman2004finding,zhang2014scalable}, an objective function based on a particular goodness of fit measure is proposed and the corresponding optimization
over partitions can be solved in any number of ways. Undoubtedly, the most widespread measure in this category is modularity Q proposed by Newman and Girvan~\cite{newman2004finding}. Modularity maximization favors putting the nodes with large number of connections inside the clusters compared to the expected connections under a random graph with the same degree sequence.

Recently, Ref.~\cite{newman2016community} showed that multiresolution modularity (Q-MR) maximization is mathematically equivalent to a special case of the DC-SBM, under a $k$-planted partition parameterization. The Q-MR algorithm works implicitly like a likelihood maximization algorithm, except that it chooses its resolution parameter, which sets the number of communities $k$, by iterating between the Q and \mbox{DC-SBM} formulations of the model. In another modularity based approach~\cite{zhang2014scalable}, the authors propose a message passing algorithm (Q-MP) by introducing a Gibbs distribution utilizing the modularity as the Hamiltonian of a spin system, and a means for model selection via minimization of the Bethe free energy. This approach enables marginalization over the ruggedness of the modularity landscape, providing a kind of complexity control not available through traditional modularity maximization.
The main issue is that to infer informative communities, some parameters of the model (inverse temperature $\beta$) need to be chosen so that the model does not enter the spin-glass phase. 
Ref.~\cite{kawamoto2016comparative} builds on this approach by proposing a generalized version of modularity message passing (Q-GMP) for model selection that infers the parameters of Boltzmann distribution (inverse temperature $\beta$) instead of just setting it to some pre-calculated value.

Spectral methods using eigen decomposition techniques can find the informative eigenvectors of an adjacency matrix or a graph Laplacian, and the embedded coordinates can be used for community detection. However, traditional spectral approaches are not appropriate in clustering sparse networks, or networks with heavy-tailed degree distributions. Recently, Ref.~\cite{Krzakala2013} proposed a spectral approach for community detection in sparse networks based on the non-backtracking matrix (S-NB), that succeeds all the way down to the detectability limit in the stochastic block model~\cite{decelle2011asymptotic}. In this setting, the number of communities $k$ is chosen by the number of real eigenvalues outside the spectral band. More recently, Ref.~\cite{saade2014spectral} proposes to choose $k$ as the number of negative eigenvalues of the Bethe Hessian matrix. Ref.~\cite{le2015estimating} proves the consistency of these approaches in dense and sparse regimes and also describes some corrections on the spectral methods of Refs.~\cite{Krzakala2013} and~\cite{saade2014spectral} (S-cBHm and S-cBHa).

There is another venue of embedding techniques used in clustering, related to feature learning in networks. Following the recent achievements in natural language processing via the skip-gram model, Ref.~\cite{perozzi2014deepwalk} develops an algorithm to encode a representation of graph vertices by modeling and then embedding a stream of rigid random walks. Ref.~\cite{grover2016node2vec} generalizes this idea and proposes an algorithm to learn continuous feature representations for nodes, which can be used in community detection and for learning which nodes have similar structure. Two attractive properties of such node-embedding approaches are their scalability and the ease with which they can be used to make predictions about edges. These methods are not included in our study as they have not yet been well explored in the context of community detection.

Traditional approaches to evaluating and controlling for model overfit, such as optimizing the bias variance tradeoff, fail in network settings because pairwise interactions violate standard independence assumptions. Because of this non-independence issue, cross-validation techniques are not theoretically well developed in the context of networks, and even simple edge-wise cross-validation can be computationally expensive. Recently, Ref.~\cite{kawamoto2016cross} showed that the leave-one-out cross-validation prediction error can be efficiently computed using belief propagation (BP) in sparse networks and thereby efficiently used for model selection. Similarly, Ref.~\cite{chen2014network} estimates the number of communities using a block-wise node-pair cross-validation method, which can be adapted to any other algorithm and model. The number of communities is chosen by validating on the testing set (minimizing the generalization error) and the technique can simultaneously chooses between SBM and DC-SBM by selecting the minimum validation error. However, it should be noted that recently Ref.~\cite{valles-catala_consistency_2017} showed that model selection techniques based on cross-validation are not always consistent with the most parsimonious model and in some cases can lead to overfitting.

Statistical methods test the number of clusters using some test statistics through a recursive hypothesis testing approach. In general, these approaches have a high computational complexity because of this outer loop. Ref.~\cite{chen2016phase} proposes an algorithm for automated model order selection (AMOS) in networks for random interconnection model (RIM) (a generalization of the SBM). The method uses a recursive spectral clustering approach, which increases the number of clusters and tests the quality of the identified clusters using some test statistics achieved by phase transition analysis. Ref.~\cite{chen2016phase} proves this approach to be reliable under certain constraints. Ref.~\cite{wang2015likelihood} proposes a likelihood ratio test (LRT-WB) statistic for the SBM or DC-SBM to choose the number of clusters $k$, and shows that when the average degree grows poly-logarithmically in $N$, the correct order of a penalty term in a regularized likelihood scheme can be derived, implying that its results are asymptotically consistent.

Ref.~\cite{bickel2016hypothesis} proposes a sequential hypothesis testing approach to choose $k$.
At each step, it tests whether to bipartition a graph or not. To this end, the authors derive and utilize the asymptotic null distribution for Erd\H{o}s-R\'enyi random graphs. This possibility originated from the fact that the distribution of the leading eigenvalue of the normalized adjacency matrix under the SBM converges to the Tracy-Widom distribution. Ref.~\cite{lei2016goodness} uses recent results in random matrix theory to generalize the approach of Ref.~\cite{bickel2016hypothesis} to find the null distribution for SBMs in a more general setting. Utilizing this null distribution and using the test statistic as the largest singular value of the residual matrix, computed by removing the estimated block model from the adjacency matrix, Ref.~\cite{lei2016goodness} proposes an algorithm to choose $k$ by testing $k=k_0$ versus $k>k_0$ sequentially for each $k_0\ge1$, which is proved to be consistent under a set of loose constraints on the number of clusters ($k=O(N^{1/6}-\tau)$ for some $\tau>0$) and the size of clusters ($\Omega(N^{5/6})$).

It is noteworthy that Ref.~\cite{ailon2015iterative} recently proved that one constraint on the sizes of inferred communities under some methods is an artifact of identifying the large and small clusters simultaneously. It goes on to show that this issue can be resolved using a technique called ``peeling,''  which first finds the larger communities and then, after removing them, finds the smaller-sized communities using appropriate thresholds. This iterative approach is similar to the superposition coding technique in coding theory and recalls the hierarchical clustering strategy introduced in Ref.~\cite{peixoto2014hierarchical} for capturing clusters with small sizes. Basically, by iteratively limiting the search space, finding an optimum solution becomes computationally more tractable. Relatedly, Ref.~\cite{chen2016statistical} shows that in planted $k$-partition model, the space of parameters of the model divides into four regions of impossible, hard, easy and simple, which are related to the regimes that algorithms based on maximum likelihood estimators can succeed theoretically and/or computationally. These results indicate that no computationally efficient parametric algorithm can find clusters if the number of clusters increase unbounded over $\Omega({\sqrt{N}})$. This fact is in strong agreement with our experimental results.

\section*{Acknowledgments}
  The authors thank Tiago Peixoto, Leto Peel, Daniel Larremore, and Martin Rosvall for helpful conversations, and acknowledge the BioFrontiers Computing Core at the University of Colorado Boulder for providing High Performance Computing resources (NIH 1S10OD012300) supported by BioFrontiers IT. The authors also thank Mark Newman, Rachel Wang, Peter Bickel, Can Le, Elizaveta Levina, Tatsuro Kawamoto, Kohei Hayashi, Pin-Yu Chen, and Etienne C\^ome for sharing software implementations. The authors thank Ellen Tucker for help with network data sets from ICON. Financial support for this research was provided in part by Grant No. IIS-1452718 (AG, AC) from the National Science Foundation.

\ifCLASSOPTIONcaptionsoff
  \newpage
\fi

\end{document}